\documentclass[preprint]{elsarticle}
\usepackage{amsmath}
\usepackage{amsfonts,amssymb}
\usepackage{bm}
\usepackage{cases}
\usepackage{subfigure}
\usepackage{booktabs}
\usepackage{array}
\usepackage{multirow}
\usepackage{listings}
\usepackage{algorithm}
\usepackage{algpseudocode}
\usepackage{marvosym}
\usepackage{graphicx} %
\usepackage{epsfig}
\usepackage{fancyhdr}
\usepackage{setspace}
\usepackage{helvet}
\usepackage{makecell}
\newcommand{\tabincell}[2]{\begin{tabular}{@{}#1@{}}#2\end{tabular}}

\usepackage{setspace}
\makeatletter
\let\ps@pprintTitle\ps@plain
\makeatother

\begin{document}
\begin{frontmatter}
\title{A panoramic aerodynamic performance prediction method for turbomachinery cascades using transformer-enhanced neural operator\tnoteref{version-note}}
\tnotetext[version-note]{Author manuscript with core data aligned to the published reference. Related publication: Chinese Journal of Aeronautics 38(7) (2025), 103473; DOI: 10.1016/j.cja.2025.103473. Figures identified in their captions are reproduced from the published article under the Creative Commons Attribution 4.0 International license (https://creativecommons.org/licenses/by/4.0/).}

\author[1]{Qineng Wang}
\author[1]{Zhendong Guo}
\author[1]{Liming Song\corref{cor1}}
\ead{songlm@xjtu.edu.cn}
\author[2,3]{Tianyuan Liu\corref{cor2}}
\ead{tianyuanliu1992@gmail.com}
\cortext[cor1]{Corresponding author}
\cortext[cor2]{Corresponding author}
\affiliation[1]{organization={Institute of Turbomachinery, Xi'an Jiaotong University},city={Xi'an},postcode={710049},country={China}}
\affiliation[2]{organization={ENN Science and Technology Development China Co., Ltd.},city={Langfang},postcode={065001},country={China}}
\affiliation[3]{organization={Hebei Key Laboratory of Compact Fusion},city={Langfang},postcode={065001},country={China}}

\begin{abstract}
To enable flexible and rapid aerodynamic performance evaluation in turbomachinery design, this paper proposes a panoramic performance prediction framework. Unlike most previous prediction models that directly predict the objective functions of interest, our approach first predicts the basic parameters of the Navier-Stokes equations, such as temperature, pressure, and density.
Utilizing these basic physical quantities, it subsequently predicts key performance parameters of the turbine stage meridian plane.
By adopting this methodology, our proposed panoramic performance prediction framework functions similarly to a CFD simulator, capable of predicting various objective of interest to the designers.
To enhance prediction accuracy, a transformer-enhanced neural operator (TNO) is introduced within this framework.
Using the Rotor 37 blades as a reference, the proposed TNO is trained to predict the performance of a transonic compressor blade in the meridian plane. The TNO can accurately predict total quantities such as isentropic efficiency, mass flow, and distributions of total pressure ratio.
Remarkably, the prediction error of TNO is observed to be smaller than that of state-of-the-art deep learning operators such as the FNO and DeepONet.
Furthermore, the TNO is applied to downstream tasks, including sensitivity analysis and optimization of various objective functions.
The results confirm that the TNO can operate almost like a CFD simulator, while reducing the computational cost of downstream tasks by four orders of magnitude.
The effectiveness and reliability of the proposed TNO for solving different kinds of downstream tasks have been well demonstrated.
\end{abstract}

\begin{keyword}
Turbomachinery cascades design; Physics Field Prediction; Deep learning; Design optimization of compressor
\end{keyword}

\end{frontmatter}

\section{Introduction}
\par
Over the last two decades, the rapidly advancing Computational Fluid Dynamics (CFD) simulation method has found extensive application in the domains of aerodynamic design~\cite{pangDatadrivenSurrogateModel2024a}, performance analysis~\cite{shenAutomaticVisibleExplainer2022}, operational forecasting, and other aerospace-related disciplines.
The application of CFD has significantly enhanced the quality, performance and reliability of aerospace equipment design (i.e. turbomachinery design).
The design cycle of turbomachinery typically includes multiple downstream tasks, such as optimization search, sensitivity analysis, uncertainty quantification.
These tasks often require a significant number of CFD calculations to obtain results for analysis.
Cost of CFD simulation per run also increases dramatically with the increase of accuracy, making it rather a challenging task to finish the design optimization task of a turbomachinery component within the allowed time and budget.
Therefore, researchers have been devoted to finding fast and accurate alternatives to CFD simulations~\cite{wuGenerativeDeepLearning2022}.
\par
Conventional surrogate models have been widely used as alternatives to time-consuming CFD simulations~\cite{guoMultiObjectiveAerodynamicOptimization2015, hanEfficientAerodynamicShape2020, buSelectingScaleFactor2022}.
Common surrogate modeling techniques include Artificial Neural Networks (ANN)~\cite{jianchangmaoArtificialNeuralNetworks1995}, polynomial regression (PR), Gaussian Process Regression (GPR), Support Vector Regression (SVR), extreme gradient boosting (XGB)~\cite{chenXGBoostScalableTree2016} and others, which significantly improving design efficiency.
However, due to limited expressive ability of conventional surrogates, the establishment of surrogate models completely neglects the rich flow field data obtained from CFD simulations and makes the predictions of surrogate models a complete black-box, lacking explanations of the underlying flow field mechanisms.
\par
To obtain more interpretable flow field prediction results, researchers have leveraged the progressive maturation of deep learning technology to predict flow fields directly, capitalizing on its end-to-end nature and powerful nonlinear represented ability.
Deep learning-based flow field prediction can rapidly attain flow field results within milliseconds~\cite{liEfficientDeepLearning2021, xiongPointCloudDeep2023}, thereby substituting CFD calculations in certain scenarios~\cite{wangProcessbasedDeepLearning2023}.
Before 2019, researchers in turbomachinery design often adopted techniques from image analysis, such as Convolutional Neural Networks (CNN), to map geometric shapes to physical fields~\cite{hennighLatNetCompressingLattice2017, zhangApplicationConvolutionalNeural2018}. CNNs were extensively used for aerodynamic flow field prediction and design, predicting quantities like velocity, pressure, and Mach number~\cite{sekarFastFlowField2019, duruDeepLearningApproach2022, kongDeepLearningApproach2023}. Accuracy improvements in CNN flow field predictions were achieved through enhanced resampling methods~\cite{wangDualconvolutionalNeuralNetwork2021, kashefiPointCloudDeepLearning2021, tangsaliGeneralizabilityConvolutionalEncoder2021}.
Graph Neural Networks (GNN) were later introduced for flow field prediction, utilizing nodes and edges to handle the unstructured discrete flow fields from CFD solutions without pixel sampling loss~\cite{liGraphNeuralNetworks2022, tangsaliGeneralizabilityConvolutionalEncoder2021, velickovicGraphAttentionNetworks2018}. Li used a GNN to predict the flow field of a radial inflow turbine~\cite{liUncertaintyQuantificationAerodynamic2023}.
In 2020, Lu introduced the concept of deep neural operator networks and designed the deep Operator Network (deepONet) framework~\cite{luLearningNonlinearOperators2021}. These operator neural networks can be trained on various discrete physical fields and predict physical quantities at any coordinate position within the domain. The Fourier Neural Operator (FNO) further enhanced this capability by producing high-resolution physical fields from low-resolution training data~\cite{liFourierNeuralOperator2021}.
These advancements have established a robust and accurate deep network architecture for flow field prediction.
\par
Using deep learning methods improves the accuracy and interpretability of predictions. However, the disadvantages of the deep learning flow field prediction method become evident when compared to traditional surrogate models. The cost of datasets and training is greatly increased, so if the trained model can only predict and simulate a specific physical field in a specific computational domain, it shows significant inefficiencies and economic drawbacks.
Therefore, this paper proposes a panoramic prediction framework to enhance the reusability of the flow field prediction model.
Under this framework, any desired physical quantity can be predicted, allowing the model to be reused for various downstream tasks.
Specifically, the model first predicts the basic physical quantities of the Navier-Stokes equation, such as temperature, pressure, density, and velocity, and then predicts any key derived performance parameters of the turbine stage meridian plane based on these basic physical quantities. Additionally, to improve the accuracy of the basic physical field flow field prediction model, this paper integrates a Transformer into the neural operator network, proposing the Transformer Neural Operator (TNO) network.
\par
This paper's primary contributions are as follows:
\par
(1)
A comprehensive panoramic prediction framework of turbomachinery components is established.
This framework offers versatile adaptability for various downstream tasks by predicting of arbitrary physical field.
\par
(2)
A transformer-enhanced operator (TNO) network by utilizing Galerkin attention~\cite{caoChooseTransformerFourier2021} is incorporate, facilitating the production of clearer and more accurate predictions.
\par
(3)
The proposed panoramic prediction method is applied to the aerodynamic design of the Rotor 37 transonic compressor.
In various downstream tasks of aerodynamic design, the panoramic frame demonstrates its flexibility and reusability, significantly reducing task time.
\par
The remainder of the paper is organized as follows.
In Section \uppercase\expandafter{\romannumeral2}, the research object, Rotor 37 blade, and its numerical simulation model will be introduced in detail.
And then, the panoramic performance prediction framework, and the TNO network, are illustrated in Section \uppercase\expandafter{\romannumeral3}.
In Section \uppercase\expandafter{\romannumeral4}, the predictive accuracy of the panoramic prediction method is analyzed and compared.
Subsequently, the obtained prediction model will be utilized for variable downstream tasks including sensitivity analysis and optimization to verify the correctness and effectiveness of the panoramic prediction method.
Finally, the conclusions are summarized in Section \uppercase\expandafter{\romannumeral5}.
\section{Problem Setup and Data Preparation}
\par
Compressors hold a pivotal position within the aerospace industry.
Notably, in the multidisciplinary domain of turbomachinery design optimization, aerodynamic design serves as the foundational and pivotal aspect of the overall design process~\cite{wangImprovedDeviationModel2024}.
\par
The main object of this research is the classic transonic compressor rotor blade, the Rotor 37 blade.
It was designed and tested by Reid and Moore at the NASA Glenn center in the 1970s~\cite{reidExperimentalStudyLow1980}.
The design parameters for this blade are shown in Table \ref{designConditionRotor37}~\cite{borettiExperimentalComputationalAnalysis2010}.
\begin{table}[htbp]
\centering
\caption{Design conditions of the Rotor 37 blade}
\begin{tabular}{cc}
\toprule
Condition name & Value \\
\midrule
equivalent rotational speed[rpm] & 17188.7 \\
number of rotor blades & 36 \\
rotor blade aspect ratio & 1.19 \\
tip clearance gap[mm] & 0.356 \\
inlet total temperature[K] & 288.15 \\
inlet total pressure[Pa] & 101325 \\
\bottomrule
\end{tabular}%
\label{designConditionRotor37}%
\end{table}%
This section offers a comprehensive overview of the methodology employed in this work for the Rotor 37 blade, including the generation of the geometry, mesh generation, CFD simulation, and data sampling strategy.
\begin{figure}[htbp]
  \centering
  \subfigure[geometry]{
  \begin{minipage}[t]{0.57\linewidth}
  \centering
  \includegraphics[width=0.9\textwidth]{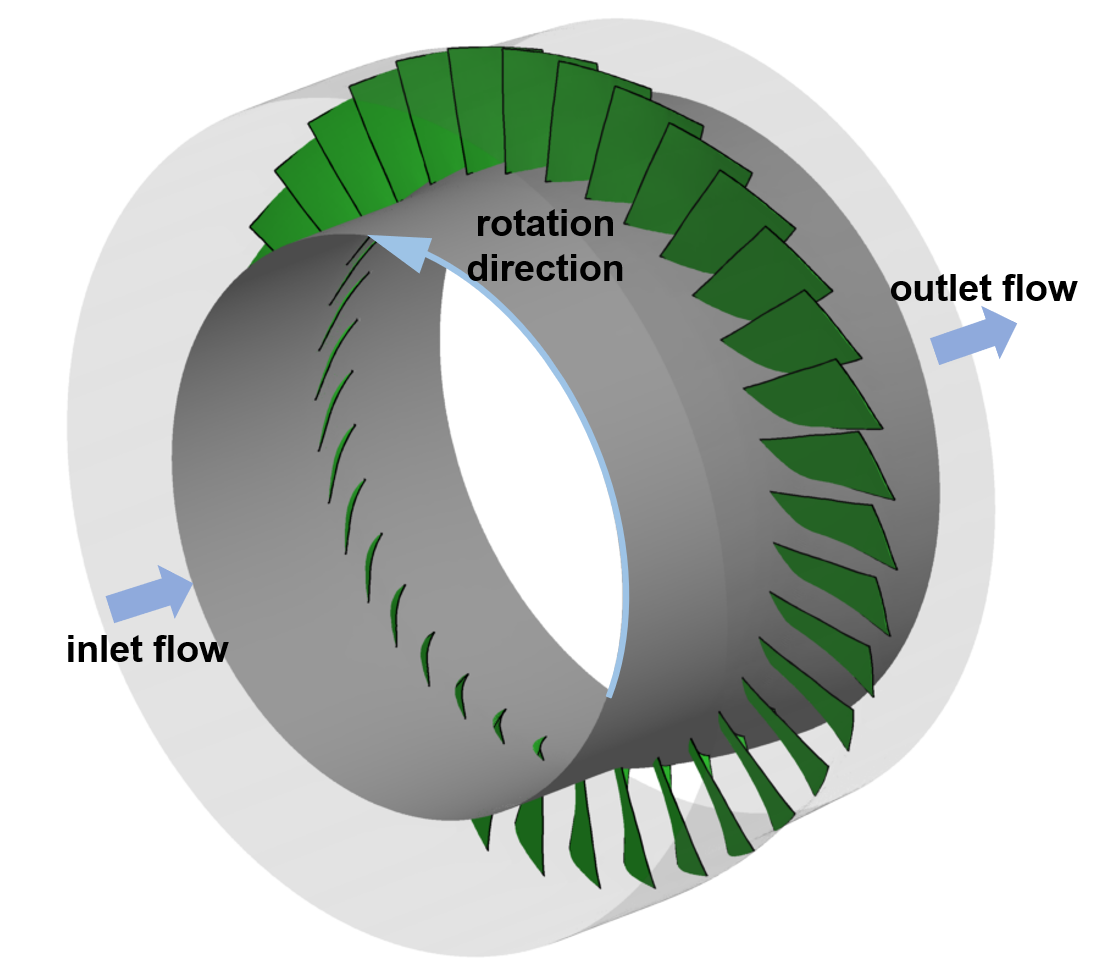}
  \end{minipage}%
  }%
  \subfigure[mesh details]{
  \begin{minipage}[t]{0.37\linewidth}
  \centering
  \includegraphics[width=0.9\textwidth]{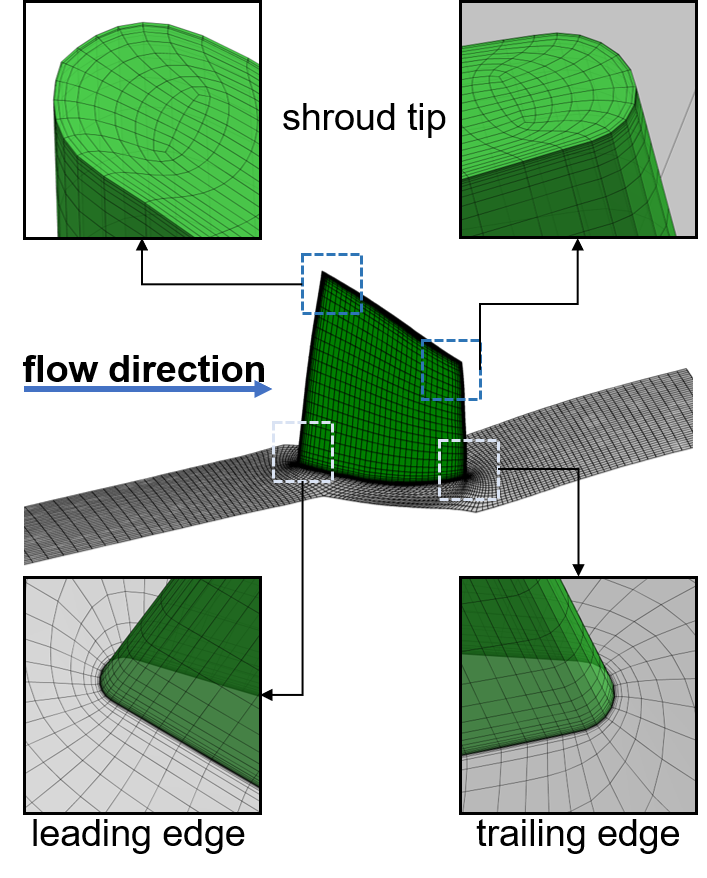}
  \end{minipage}
  }%
  \centering
  \caption{The geometry and mesh illustration of Rotor 37 blade}
  \label{rotor37_geom}
  \end{figure}
\subsection{Geometric generation}
\par
Figure \ref{rotor37_geom}(a) illustrates the geometric schematic of the Rotor 37 blade, encompassing the blade, hub, and shroud.
Under the assumption of steady flow conditions, complete flow field information can be obtained by computing a single flow passage of one blade.
The parameterization process of the three-dimensional blade geometry is accomplished using a well-established in-house code.
The Rotor 37 has a total of 28 design variables, which are listed in Table \ref{Rotor37Variable}.
The suction surface is selected for modification because of its influence on flow separation and aerodynamic performance.
Five uniformly spaced spanwise sections are used, with five control points defining each section profile through non-uniform rational B-splines (NURBS).
The first 25 variables specify control-point displacements normal to the profile; their ranges are expressed relative to the axial chord $C_{\mathrm{axial}}$.
Upon completing the parameterization of all section profiles, a single circumferential translation parameter $x_{26}$ is chosen for the middle section, while two axial translation parameters $x_{27}$ and $x_{28}$ are selected for adjusting the stacking line in 3D space in the tip and middle sections.
For a more detailed description of the parameterization methodology, please refer to this work~\cite{wangKTEGOKnowledgeTransfer2022}.
\begin{table}[htbp]
\centering
\small
\caption{Design variables of the Rotor 37 blade. Displacement ranges are multiples of the axial chord $C_{\mathrm{axial}}$.}
\setlength{\tabcolsep}{3pt}
\begin{tabular}{>{\raggedright\arraybackslash}p{0.30\linewidth}p{0.40\linewidth}p{0.23\linewidth}}
\toprule
Geometric definition & Variable index & Range/$C_{\mathrm{axial}}$ \\
\midrule
Leading edge & $x_1,x_6,x_{11},x_{16},x_{21}$ & $[-0.6\%,0.6\%]$ \\
Near-leading edge & $x_2,x_7,x_{12},x_{17},x_{22}$ & $[-1.2\%,1.2\%]$ \\
Intermediate & $x_3,x_8,x_{13},x_{18},x_{23}$ & $[-1.5\%,1.5\%]$ \\
Near-trailing edge & $x_4,x_9,x_{14},x_{19},x_{24}$ & $[-1.2\%,1.2\%]$ \\
Trailing edge & $x_5,x_{10},x_{15},x_{20},x_{25}$ & $[-0.6\%,0.6\%]$ \\
Circumferential translation & $x_{26}$ & $[-15\%,15\%]$ \\
Axial translation & $x_{27},x_{28}$ & $[-10\%,10\%]$ \\
\bottomrule
\end{tabular}
\label{Rotor37Variable}
\end{table}%
\subsection{Numerical simulation and validation}
After generating the geometry, this study employed a commercial CFD software to obtain the three-dimensional flow field information of the predicted object.
The mesh generation was accomplished using Numeca Autogrid5, employing an H-O-I grid topology.
The near-wall first layer grid thickness was set to $3 \times 10^{-6}$ m.
The structured mesh used for optimization in this study consisted of approximately $4 \times {10^{ 6}}$ nodes.
The geometry of single flow channel and mesh details of the reference design are presented in Fig. \ref{rotor37_geom}(b). The grid-independence comparison is provided in Fig. \ref{grid_independence} in the Appendix.
In the calculation, the Spalart-Allmaras turbulence model is used with adiabatic smooth walls condition.
The Reynolds-averaged Navier-Stokes equations are solved by using the commercial software NUMECA FINE/TURBO.
To be in accordance with the previous literature~\cite{borettiExperimentalComputationalAnalysis2010}, uniform total pressure and temperature are imposed at the inlet boundary, and an averaged static pressure is imposed at the outlet as a constant of 115000 Pa, corresponding to a relative mass flow rate of 99\% for the Rotor 37 blade.
With the CPU Intel(R) i5-9400F@2.9GHz, the calculation time of a single sample is about 800s.
\begin{figure}[htbp]
\centering
\subfigure[efficiency-mass flow]{
\begin{minipage}[t]{0.48\linewidth}
\centering
\includegraphics[width=1\textwidth]{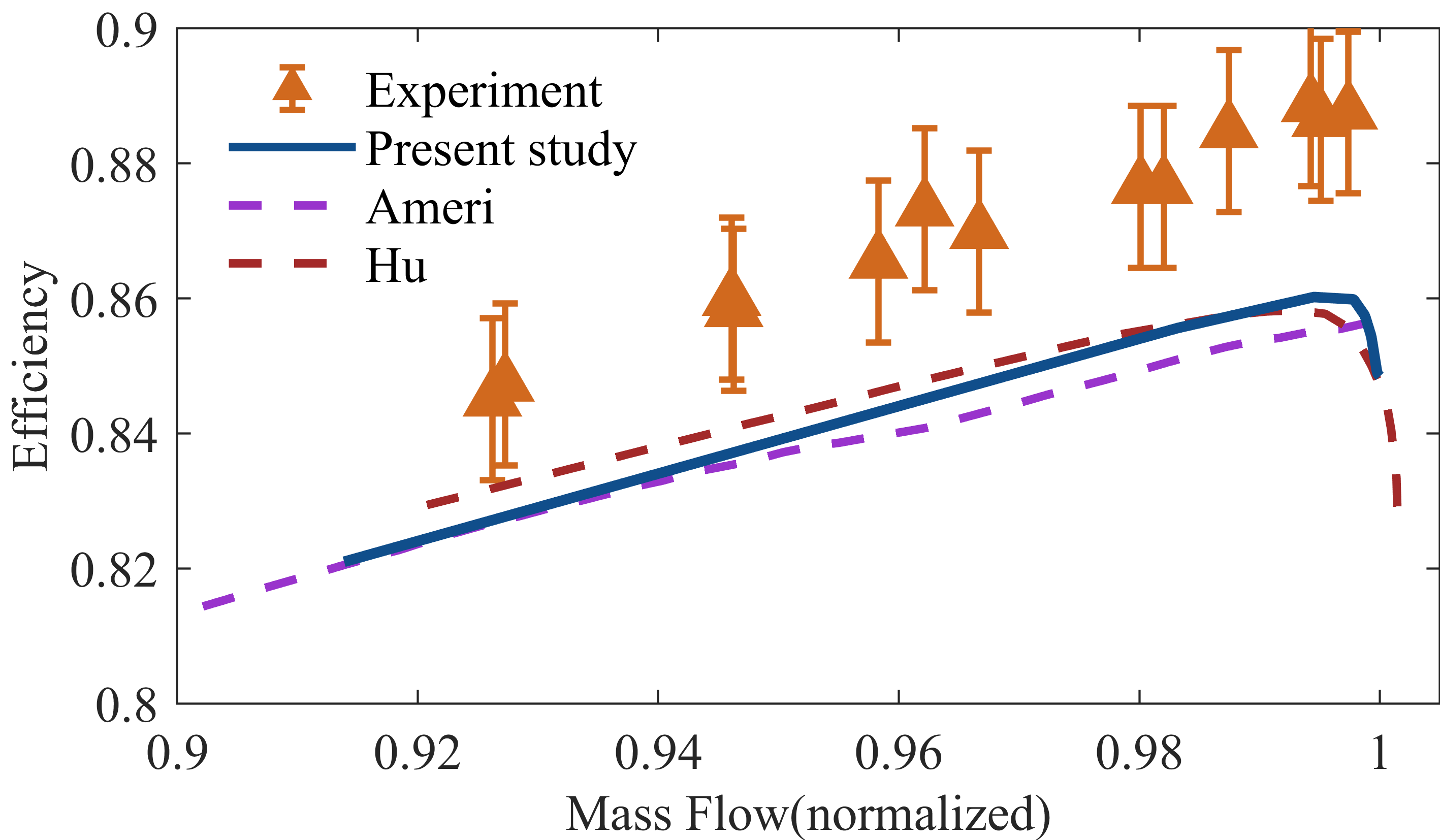}
\end{minipage}%
}%
\subfigure[pressure ratio-mass flow]{
\begin{minipage}[t]{0.48\linewidth}
\centering
\includegraphics[width=1\textwidth]{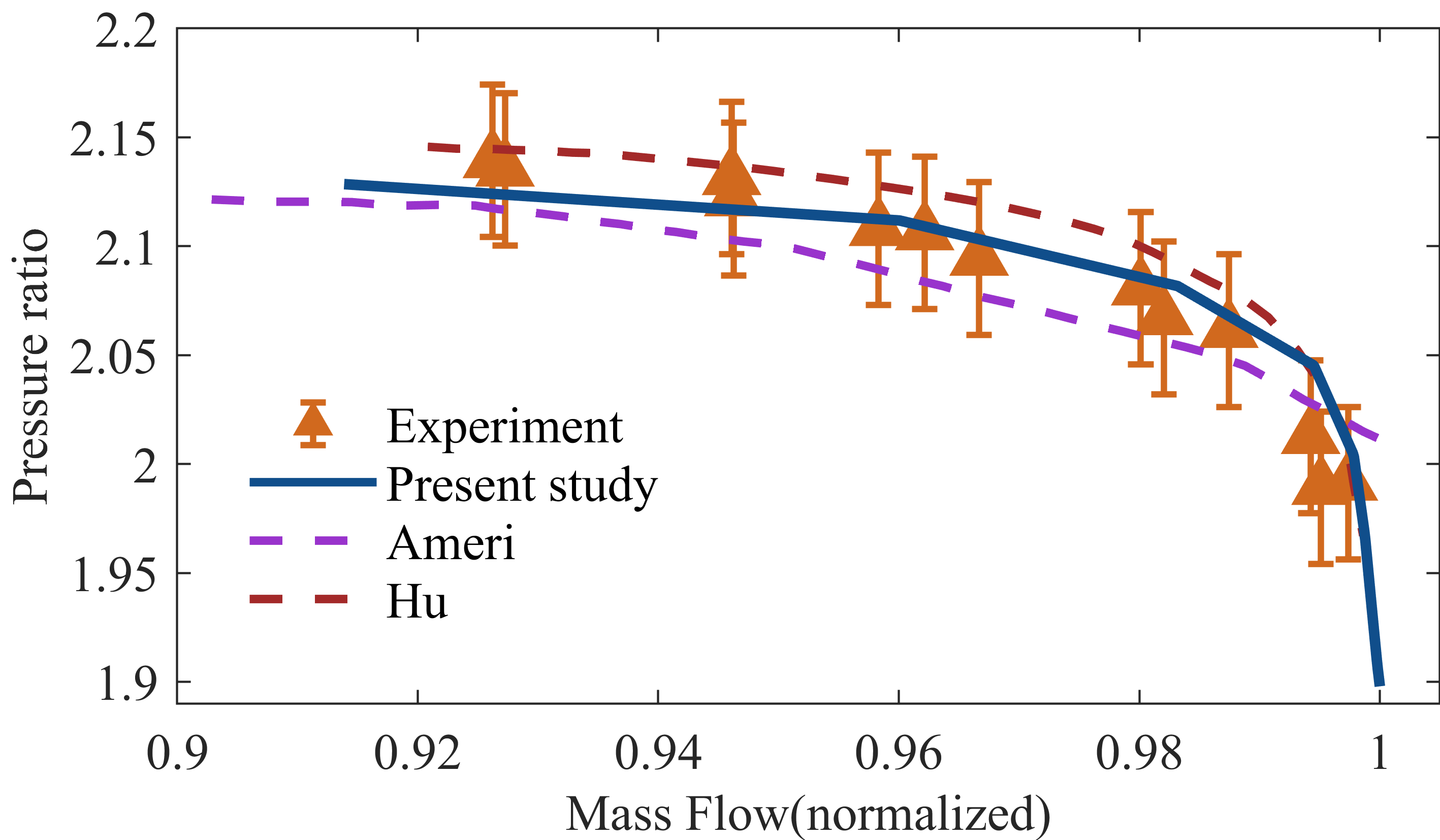}
\end{minipage}
}%
\centering
\caption{The comparison of feature curves of CFD simulation and experiment result}
\label{Rotor37EXP}
\end{figure}
Figure \ref{Rotor37EXP} illustrates the comparison between the compressor characteristic curves obtained from the present study's CFD simulations and experimental data.
It is evident that the CFD results are in excellent agreement with the experimental data for the pressure ratio curves as shown in Fig. \ref{Rotor37EXP}(b).
Although the average efficiency curves exhibit a deviation of approximately 2\% from the experimental data~\cite{reidExperimentalStudyLow1980}, this observation aligns with findings from other researchers~\cite{ameriNASAROTOR372009,huApplicationSupportVector2021}.
These results from both curves verify the accuracy of the CFD calculation method.
\subsection{Data preparation}
\par
In the process of data preparation, the Latin Hypercube Sampling (LHS) method was employed to gather 2900 samples in the design space for the 28 geometric design variables $\mathbf{x}$.
2900 data cases contain the design variables $\mathbf{x}$ and the flow fields $\mathbf{f}$ of the Rotor 37 blade are collected as $\{\mathbf{x}, \mathbf{f}\} \in \mathcal{D}$.
The prepared data will be utilized to train and validate the deep operator networks.
\par
\begin{figure}[ht]
\begin{center}
\includegraphics[scale=0.45, trim = 0 0 0 0]{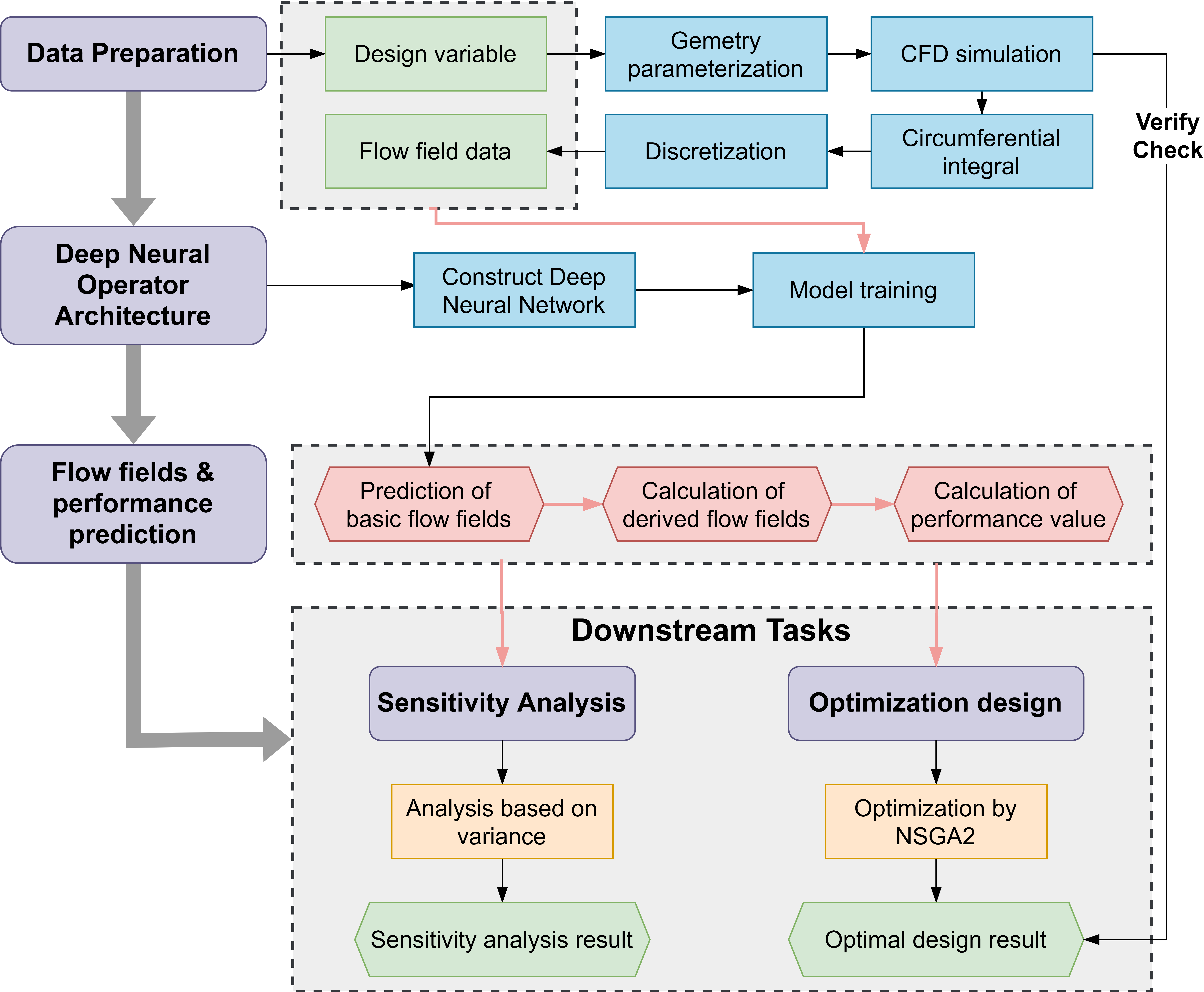}
\end{center}
\caption{The work flow of this study}
\label{overall_workflow}
\end{figure}

Figure \ref{overall_workflow} illustrates the comprehensive workflow of this paper.
This paper first completes the preparation of training data, and subsequently establishes a panoramic performance prediction network based on the collected data, evaluating its accuracy in predicting flow field and performance.
Moreover, the developed network enables near-real-time prediction of the flow fields and performance metrics for a specific blade geometry, facilitating a wide range of downstream tasks to assess the reusability of the panoramic framework.
\vspace{-0.7cm}
\section{Proposed Method}
\par
In aerodynamic shape design, designers aim to model the relationships between geometric design variables ($\mathbf{x}$), state parameters ($\boldsymbol{\alpha}$), flow field ($\mathbf{f}$), and performance metrics ($\boldsymbol{\Psi}$).
The physical quantity $\mathbf{f}(\mathbf{A})$ represents values defined within the coordinates $\mathbf{A}$ of the computation domain $\boldsymbol{\Omega}$.
To achieve the panoramic framework's mapping process as $(\mathbf{x},\boldsymbol{\alpha}) \mapsto \mathbf{f} \mapsto \boldsymbol{\Psi}$ for flow fields and performance in a fast and cost-effective way, a basic field predictor $\hat{\mathcal{F}}:(\mathbf{x},\boldsymbol{\alpha})\mapsto \mathbf{f_s}$ and a panoramic predictor $\hat{\mathcal{Y}}:(\mathbf{f}\mapsto \boldsymbol{\Psi})$ are needed.
Thus, this section first introduces a new operator network (TNO).
TNO is designed based on the characteristics of multiple physical fields governed by the Navier-Stokes equations to achieve efficient prediction of basic physical fields.
Following this, a panoramic prediction framework tailored to the characteristics of turbomachinery design is established.
Various performance parameters are then derived and calculated from the predicted basic fields and used in downstream tasks.
\subsection{Transformer-based Neural Operator architecture}
\par
The accuracy of flow field prediction relies on the architecture of the neural operator network, which is the key element for achieving efficient panoramic predictions.
To this end, this paper develops a new Transformer-enhanced Neural Operator (TNO).
Its architecture fully considers the characteristics of multi-physics field data, effectively improving accuracy.
\par
The deep neural operator is a neural network model developed based on the Universal Approximation Theorem (UAT)~\cite{tianpingchenUniversalApproximationNonlinear1995}.
Lu has proposed a framework for neural operator networks~\cite{luLearningNonlinearOperators2021}, comprising
(\romannumeral 1) A branch network that encodes the input function.
(\romannumeral 2) A trunk network that encodes the query locations.
The proposed TNO network in this article also uses this branch-trunk framework.
\par
\begin{figure}[!ht]
\begin{center}
\includegraphics[width=1\textwidth]{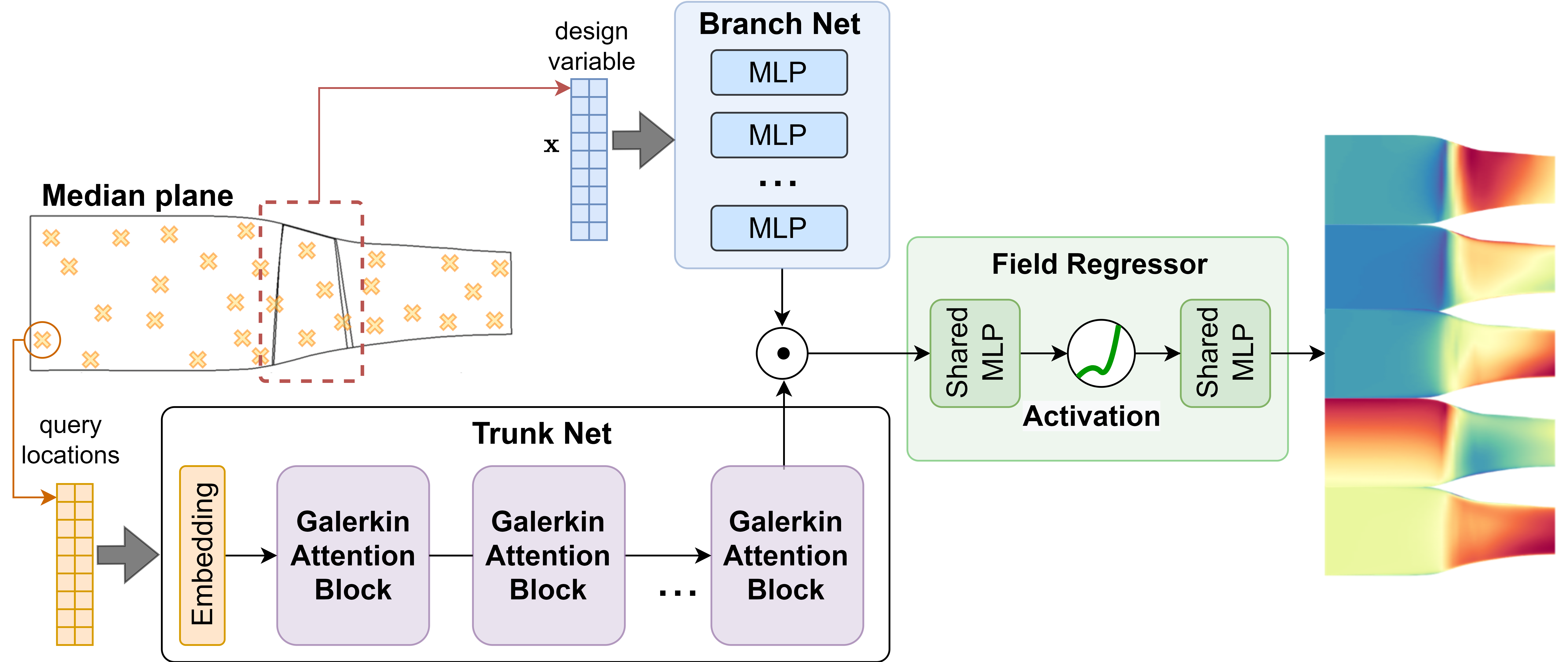}
\end{center}
\caption{The network structure of the TNO network}
\label{proposed_network}
\end{figure}

Figure \ref{proposed_network} illustrates the network architecture of the TNO, which can be expressed as:
\begin{equation}\label{TNO_structure}
\hat{\mathcal{F}}\left(\mathbf{x}\right)(r, z)=
\hat{\mathcal{S}}\left( \{
\underbrace{
  \hat{\mathcal{B}}_k\left(x ; \theta_b\right)}_{\text {branch}}
\underbrace{
  \hat{\mathcal{T}}_k\left(r, z ; \theta_t\right)}_{\text {trunk}}
|(k=1,\cdots,p)
\}
; \theta_s\right)
\end{equation}
Where, $\hat{\mathcal{B}}(\cdot)$, $\hat{\mathcal{T}}(\cdot)$, and $\hat{\mathcal{S}}(\cdot)$ denote the branch network, trunk network and shared output network respectively;
$\theta_{b}$, $\theta_{t}$, and $\theta_{s}$, denote the hyperparameters that need to be adjusted in the above three networks;
$p$ denote the dimension of the feature vector $\mathbf{B}$ and $\mathbf{T}$.
Notably, the design variables $\mathbf{x}$ and coordinates $(r,z)$ in the panoramic frame are directly utilized to describe the network input.
\par
\begin{figure}[!ht]
\begin{center}
\includegraphics[width=0.6\textwidth]{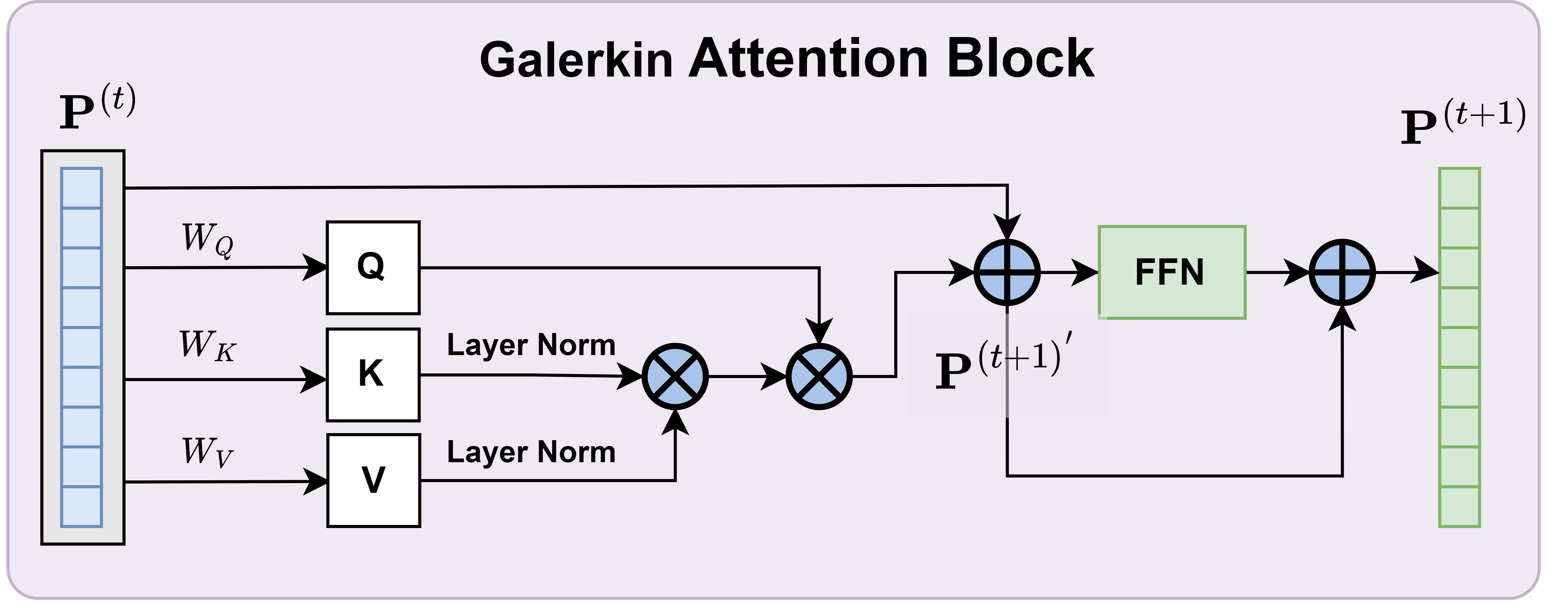}
\end{center}
\caption{The structure diagram of Galerkin attention block}
\label{galerkin_network}
\end{figure}

Specifically, the branch network encodes the design variable $\mathbf{x}$ as a branch feature $\mathbf{B}$ using a series of stacked Multilayer Perceptron(MLP) networks.
The trunk network is redesigned to accommodate the specific characteristics of flow field information, enabling unstructured input and higher predictive accuracy.
It encodes a series of query locations $(r_{j},z_{j})\quad(j=1,\cdots,M)$ into trunk feature $\mathbf{T}$ with multiple self-attention blocks in series.
Denoting the input and output of these blocks as $P^{(t)}$, the entire trunk network $\hat{\mathcal{T}}(r,z)=\mathbf{T}$ can be represented as multiple Galerkin attention blocks like:
\begin{equation}\label{Galerkin-attention-block}
\begin{split}
&\mathbf{P}^{(0)}=E\left(r,z\right),
\qquad
\mathbf{P}^{(t+1)}=\mathcal{G}\left(P^{(t)}\right), \qquad
\mathbf{T} = \mathbf{P}^{(t_m)} \\
&\mathbf{P}^{(t+1)^{'}}=\mathbf{P}^{(t)}+\text{Attn}(\mathbf{P}^{(t)}),\qquad
\mathbf{P}^{(t+1)}=\mathbf{P}^{(t+1)^{'}}+\text{FFN}(\mathbf{P}^{(t+1)^{'}})
\end{split}
\end{equation}
Where, $E(\cdot)$ denotes the embedding block.
$\mathcal{G}(\cdot)$ denotes a Galerkin attention blocks as shown in Fig. \ref{galerkin_network}.
$\text{FFN}(\cdot)$ denote a Feed Forward Networks(FFN)
\par
This paper employs the Galerkin-based simplified self-attention technique~\cite{caoChooseTransformerFourier2021} in place of the conventional vanilla attention mechanism~\cite{vaswaniAttentionAllYou2017}.
This modification simplifies the ``softmax'' computation, thereby enhancing calculation speed.
Moreover, focus on the $\text{Attn}(\cdot)$ in the Equation(\ref{Galerkin-attention-block}), the intermediate result of this step is expressed as $\mathbf{Z}=\text{Attn}(\mathbf{P}^{(t)})=\text{Attn}(\mathbf{Q},\mathbf{K},\mathbf{V})$.
Here, $\mathbf{Q}$, $\mathbf{K}$ and $\mathbf{V}$ denote the matrix of query, keyword and value respectively, which is obtained by multiplying the coefficient with the input value as $\mathbf{Q}=\mathbf{P}^{(t)}\mathbf{W}^Q$.
And for the intermediate result $\mathbf{Z}$, its $i^{\text{th}}$ element of the $j^{\text{th}}$ column vector is record as $\left(\mathbf{z}^{j}\right)_i$, and the value of
$\left(\mathbf{z}^{j}\right)_i$ is defined as follows:
\begin{equation}\label{Galerkin-attention}
\left(\mathbf{z}^{j}\right)_i=\sum_{h=1}^d \frac{\left(\mathbf{k}^h \cdot \mathbf{v}^j\right)}{M}\left(\mathbf{q}^h\right)_i \approx \sum_{h=1}^d\left(\int_{\Omega}\left(k_h(\xi) v_j(\xi)\right) d \xi\right) q_h\left(x_i\right)\\
\end{equation}
Where, $\Omega$ denotes the spatial domain that is discretized by $M$ points.
As shown in the Fig. \ref{galerkin_network}, the Equation (\ref{Galerkin-attention}) can also be expressed as $\mathbf{Z}=\mathbf{Q}\left(\widetilde{\mathbf{K}}^T \widetilde{\mathbf{V}}\right)/m$.
Where $\widetilde{\cdot}$ denotes a trainable layer normalization~\cite{baLayerNormalization2016}.
\par
After obtain $\mathbf{B}$ and $\mathbf{T}$ from the branch net $\hat{\mathcal{B}}(\cdot)$ and trunk net $\hat{\mathcal{T}}(\cdot)$ respectively, the encoded information of geometric variables is effectively applied to the global flow field with the operation of dot product.
Then the TNO network connects a shared output network $\hat{\mathcal{S}}(\cdot)$ to further transform the dot producting of $\mathbf{B}$ and $\mathbf{T}$ by cross multiplication before output results, which facilitates simultaneous processing of multiple physical field information in panoramic prediction as $\hat{\mathcal{S}}(\langle\mathbf{B},\mathbf{T} \rangle )$.
\subsection{Panoramic performance calculation framework}
\par
Particularly in the design of turbomachinery, designers commonly prioritize the assessment of flow performance, which evaluates the overall aerodynamic state between two flow surfaces, and can be expressed as:
\begin{equation}
  \begin{split}
  {{\bf{\psi }}_j} &= {{\mathcal Y}_j}({\bar{\bar{\mathbf{f}}}_1}({z_1}),{\bar{\bar{\mathbf{f}}}_2}({z_2}))\\
  \bar{\bar{\mathbf{f}}}(z) &= \frac{{\int_{{r_h}}^{{r_s}} {\bar f} (z,r) \cdot \overline {\rho {v_m}} (z,r)dr}}{{\int_{{r_h}}^{{r_s}} {\overline {\rho {v_m}} } (z,r)dr}}\\
  \bar{\mathbf{f}}(z,r) &= \frac{{\int_{{\theta _s}}^{{\theta _p}} f (z,r,\theta ) \cdot \rho {v_m}(z,r,\theta )r d\theta }}{{\int_{{\theta _s}}^{{\theta _p}} \rho  {v_m}(z,r,\theta )r d\theta }}
  \end{split}
  \end{equation}
where $\mathbf{A} = (\theta,r,z)$ denote the direction of rotation, spanwise, and stream wise, respectively;
$\theta_p$ and $\theta_s$ denote the coordinates on the periodic boundaries of the flow passage, and when the integration path intersects the blade surface, $\theta_p$ and $\theta_s$ correspond to the edges of the blade;
$r_h$ and $r_s$ denote the $r$ coordinates of the hub and shroud.
$\rho v_m$ denotes the local mass flow rate.
\par
\begin{figure}[!ht]
\begin{center}
\includegraphics[width=0.90\textwidth]{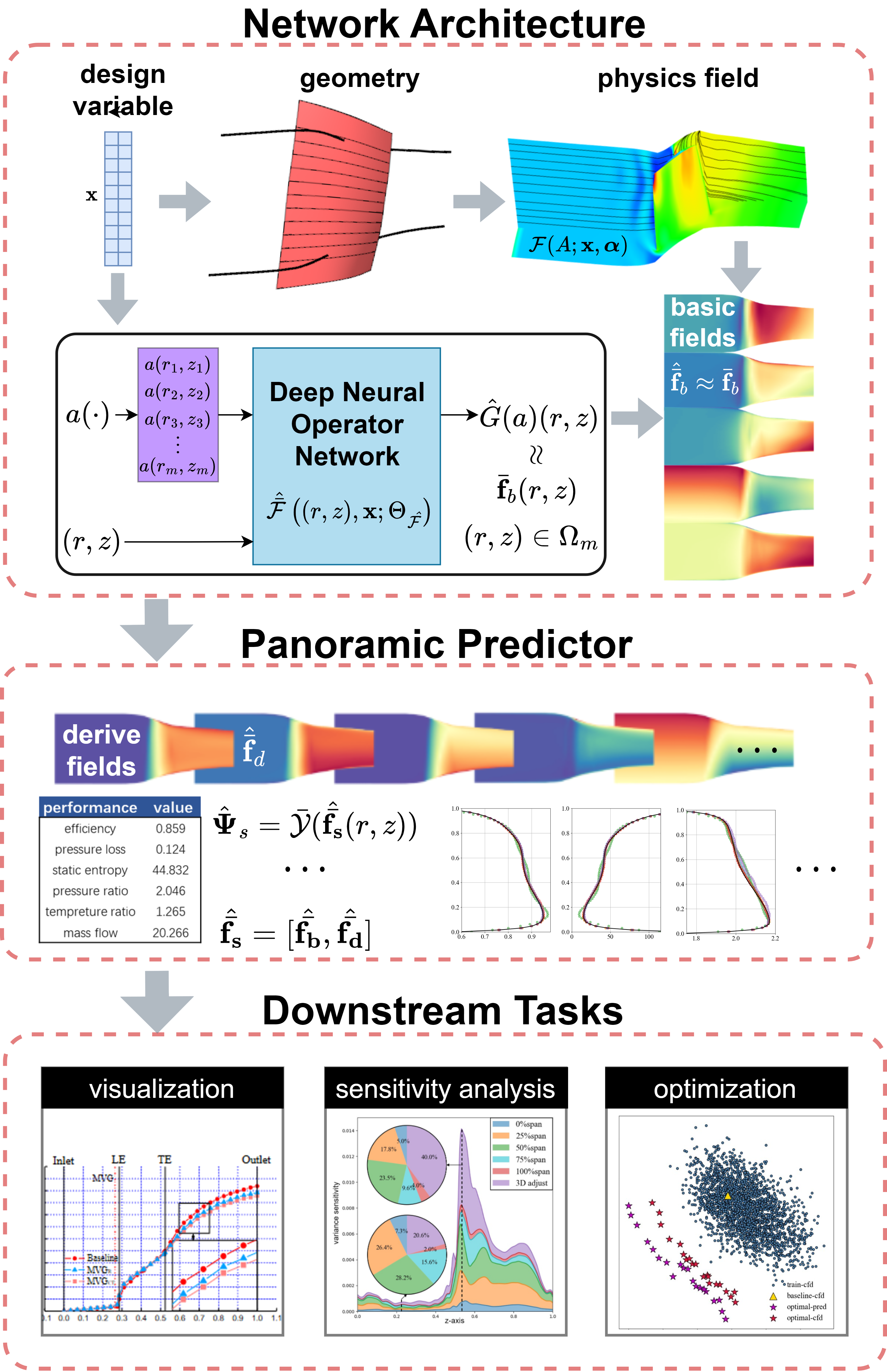}
\end{center}
\caption{The panoramic flow field and performance prediction framework}
\label{framework}
\end{figure}

The computation of averaged physical quantity in evaluating global performance metrics can be decomposed into circumferential integration and spanwise integration.
By executing these two processes before and after flow field prediction respectively, we can establish the panoramic flow field and performance prediction framework depicted in Fig.\ref{framework}.
The complete panoramic flow field and performance prediction framework, can be segmented into three components: network structure, panoramic predictor, and downstream task.

\subsection{Training strategy}
\par
This paper selects the following five quantities as the basic fields:
static pressure $p$,
static temperature $t$,
absolute velocity $v$,
relative velocity $w$,
and mass flow rate $\rho v_m$.
Following weighted integration, these five physical quantities are transformed into $\bar{\mathbf{f}}_b = [\bar{p}, \bar{t}, \bar{v^2}, \bar{w^2}, \bar{{\rho v_m}}]$, where $v$ and $w$ are squared for computational convenience.
The panoramic performance calculation is based on these basic physical fields.
As all physical fields used in network training are integrated, the single horizontal superscript of the physical field symbol will be omitted in the following paper for clarity, when there is no ambiguity.
Then, each weighted meridian plane $\bar{\mathbf{f}}_b$ contained all 5 basic fields is discretized into $M$ points, denoted as $\mathcal{P}_\text{d} =\{(r_{j},z_{j})|j=1,\cdots,M\} $, here $M=4096$.
Then, the transformed dataset is divided into training set with 2500 samples, and validation set with 400 samples.
In order to maintain consistent weighting of the five physical fields throughout the prediction process, all data are normalized using mean and standard deviation normalization prior to training.
The training process can be represented by $\Theta_{\mathcal{F}}^*=\underset{\Theta_{\mathcal{\hat{F}}}}{\arg \min }\left\{\mathbb{E}_{\{\mathbf{x}, {\mathbf{f}_b}\}-\mathcal{D}_\text{train}}\left(\mathcal{L}_{{\mathbf{f}}}\right)\right\}$ , where $\mathcal{L}$ denotes the Mean Squared Error (MSE) loss as $\mathcal{L}_{{\mathrm{f}}}(\hat{\mathbf{f}}_b, \mathbf{f}_b)=\|{\mathbf{f}}_b-\hat{\mathbf{f}}_b\|_{\boldsymbol{\Omega}_m}$.
\par
The Adam optimizer is used to optimize the aforementioned parameters.
The regularization coefficient is set to $10^{-6}$.
In this paper, all networks are trained for 900 epochs on the training set.
In the first 800 epochs, the learning rate is set to $0.001$, while in the last 100 epochs, the learning rate is reduced to $0.0001$.
The training processes for the all networks were performed using the NVIDIA GeForce-3080Ti GPU.

\section{Results and discussion}
\par
The proposed panoramic prediction framework is discussed in detail above.
In Section 4, the effectiveness and efficiency of the framework including the TNO network will be demonstrated through the completion of an aerodynamic design task involving Rotor 37 blade.
\subsection{ The prediction accuracy of the panoramic prediction framework}
For comparison, several other neural networks, MLP, UNet, deepONet, and FNO, are all utilized to perform the same prediction task.
The UNet is a widely adopted network for flow field prediction, whereas deepONet and FNO are considered classical deep neural operators.
This section will compare the prediction accuracy of flow field and performance parameters for the proposed TNO network.
\subsubsection{ Basic fields and derived fields prediction}
\par
\begin{figure}[!ht]
  \begin{center}
    \includegraphics[width=0.8\textwidth,trim = 0 0cm 0 0cm, clip]{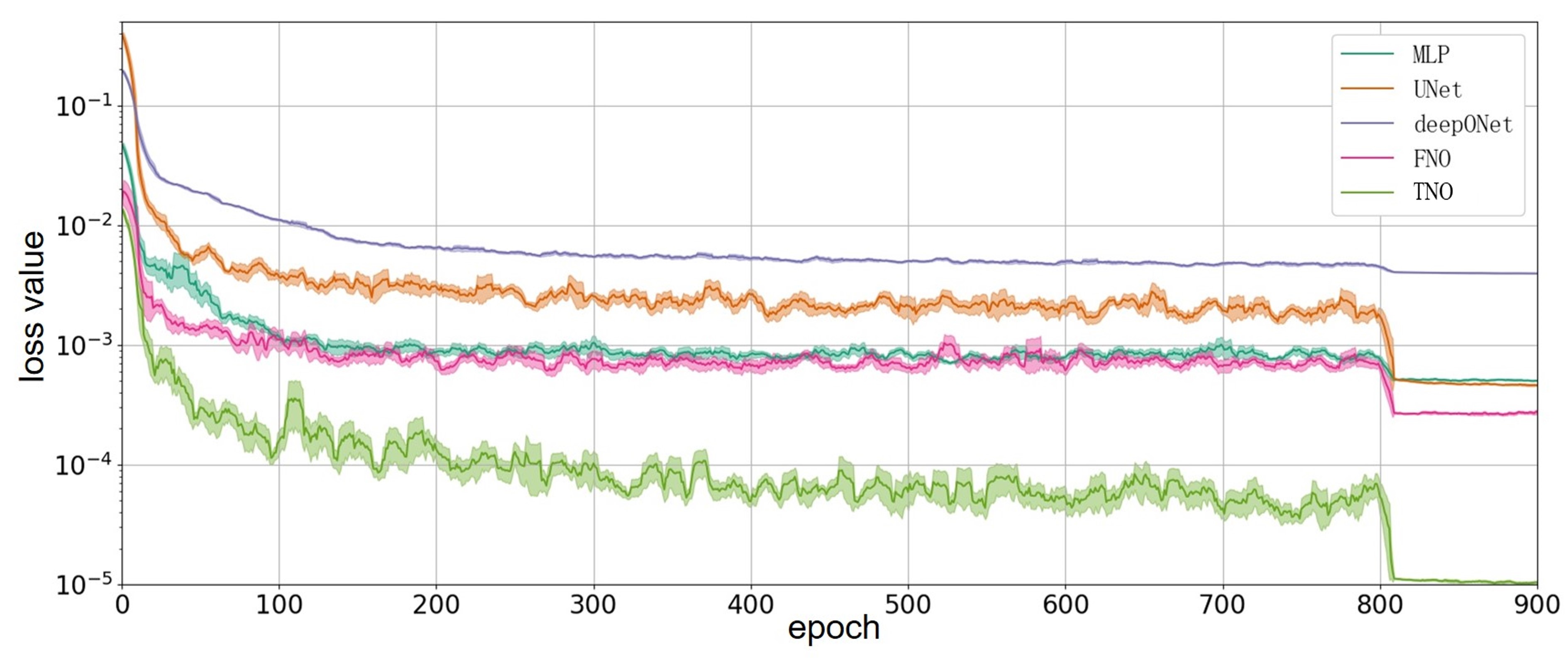}
  \end{center}
  \caption{The convergence curves of loss functions of different networks}
  \label{lossFuncCurve}
  \end{figure}

Initially, a comparative analysis is conducted to assess the accuracy of predicting basic physical field information using various neural operator networks.
Table \ref{times} displays the parameter number and forward memory size, training and flow field prediction times of 5 distinct deep neural networks.
Remarkably, the TNO's prediction time is approximately $0.9$ms, which is 5 orders of magnitude lower than that of the CFD simulation.
The specific parameters and structural details of the comparison network are shown in the Appendix.
\begin{table}[htbp]
\centering
\caption{The comparison of different network model establishment and training details}
\begin{tabular}{cccccc}
\toprule
Network
& MLP
& UNet\cite{ronnebergerUNetConvolutionalNetworks2015}
& deepONet\cite{luLearningNonlinearOperators2021}
& FNO\cite{liFourierNeuralOperator2021}
& TNO  \\ \midrule
\makecell[c]{parameter number\\ $[\times10^3]$} & 5731  & 3418  & 302  & 436  & 154   \\
\makecell[c]{forward memory\\ $[\text{MB}]$} & 768.0  & 624.5  & 822.5  & 558.9  &  147.13 \\
\makecell[c]{training time\\ $[\text{min}]$} & 19.98  & 114.0  & 52.9 & 106.5  & 40.5   \\
\makecell[c]{predicting time\\ $[\text{ms}]$} & 2.198  & 5.075  & 0.970  & 1.448  & 0.915   \\
\bottomrule
\end{tabular}
\label{times}%
\end{table}
Figure \ref{lossFuncCurve} displays the convergence curves of the loss function on the validation set in the training process.
The TNO network exhibits a significantly smaller loss function compared to the other models, approaching a value of $1\times10^{-5}$.
Table \ref{related_error} exhibits the prediction accuracy of the networks for the five basic physical fields upon completion of training, The definition of the fields related error (FRE) is $e_\text{FRE} = \frac{1}{N_\text{valid} M} \sum_{i=1}^{N_\text{valid}}\sum_{j=1}^{M} {\sqrt{(\mathbf{f}_{b,ij}-\hat{\mathbf{f}}_{b,ij})^2}/{|\mathbf{f}_{b,ij}|}}$.
\begin{table}[htbp]
\centering
\caption{The related error of the predicted basic physical fields}
\begin{tabular}{ccccccc}
\toprule
& $p$ & $t$ & $v^2$ & $w^2$ & $\rho v_{m}$ & average  \\
\midrule
MLP & 0.624\% & 0.245\% & 1.323\% & 0.954\% & 0.956\% & 0.820\%  \\
UNet & 0.723\% & 0.229\% & 1.342\% & 0.852\% & 0.811\% & 0.792\%  \\
deepONet & 1.244\% & 0.609\% & 2.894\% & 2.436\% & 2.257\% & 1.888\%  \\
FNO & 0.472\% & 0.166\% & 1.054\% & 0.624\% & 0.546\% & 0.572\%  \\
TNO & \bf{0.060}\% &	\bf{0.024}\% &	\bf{0.153}\% &	\bf{0.087}\% &	\bf{0.091}\% &	\bf{0.083}\% \\
\bottomrule
\end{tabular}%
\label{related_error}%
\end{table}%
Comparing the related error of the networks, it is evident that the TNO network has the highest prediction accuracy.
It exhibits a significant advantage by achieving an average error reduction of 85.5\% across the FNO network.
\par
\begin{figure}[htbp]
\begin{center}
\includegraphics[scale=0.4, trim = 3cm 1cm 3cm 1cm,clip]{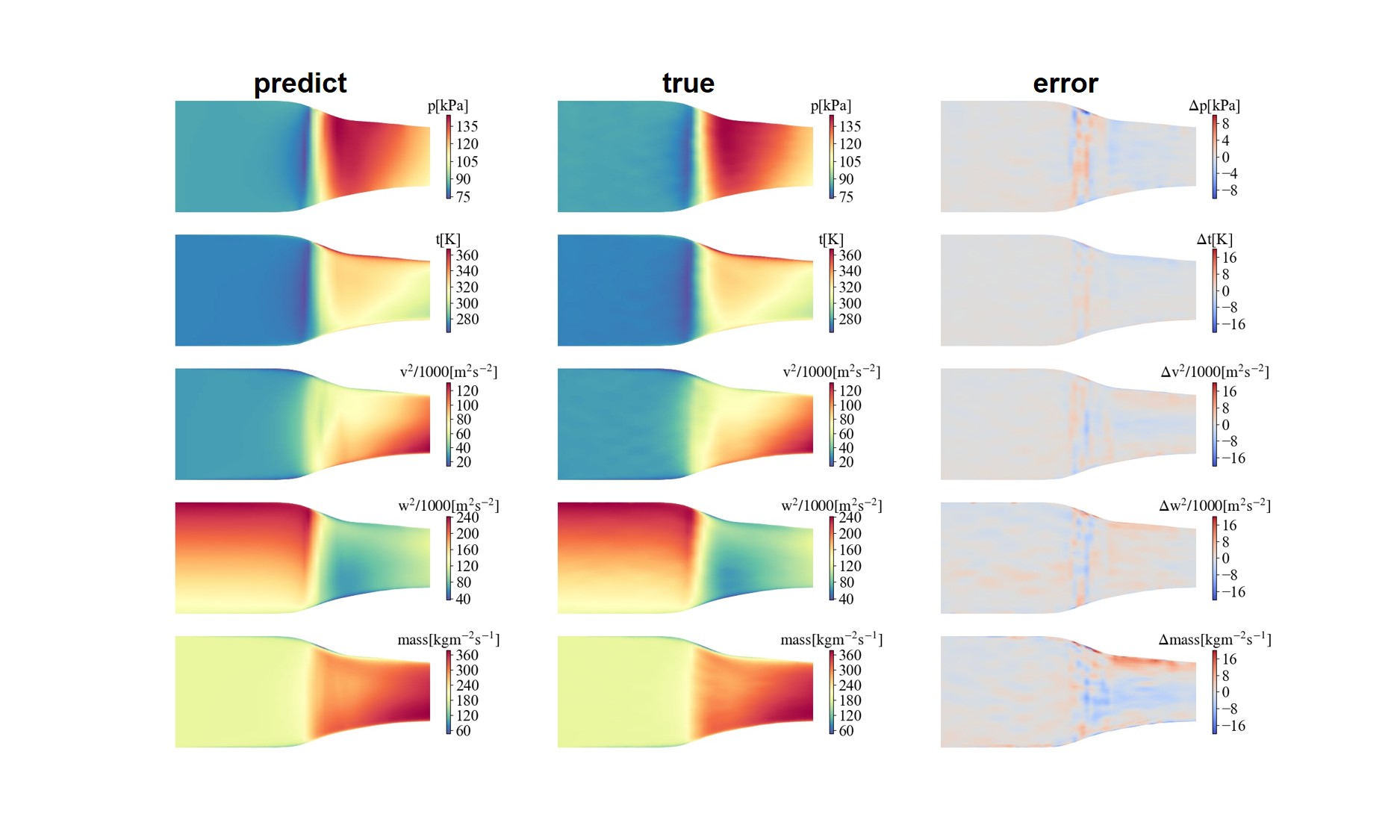}
\end{center}
\caption{Comparison of TNO network predicted and CFD results of basic flow fields for the case valid-A}
\label{fundamental_field_A}
\end{figure}

Three cases (labeled as valid-A, B, and C) are chosen randomly from the validation dataset for further illustration. The case with the worst TNO prediction is also selected from the validation dataset and labeled valid-D.
Figure \ref{fundamental_field_A} depicts the comparison between the TNO predicted and CFD simulation result of case valid-A, demonstrating precise predictions across all five basic physical fields.
And the illustration of cases valid-B and valid-C can be found in Appendix.
\par
\begin{figure}[htbp]
\begin{center}
\includegraphics[scale=0.4, trim = 3cm 1cm 3cm 1cm,clip]{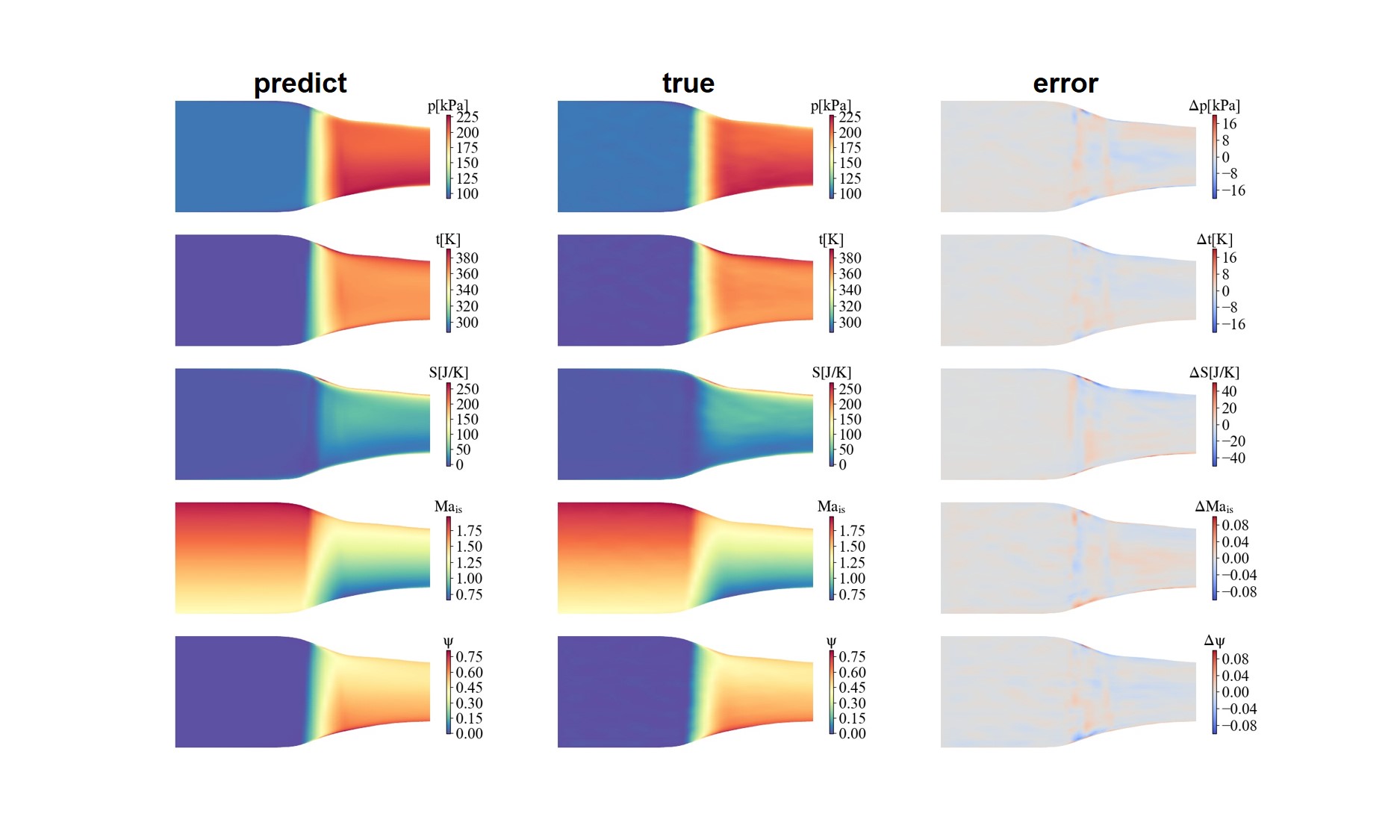}
\end{center}
\caption{Compare of predicted and CFD results of derived flow fields for case valid-A}
\label{derived_field_A}
\end{figure}

There are a series of derived fields that designers are very concerned about in aerodynamic design, and these fields can be expressed by the above five basic physical quantities as shown in Table \ref{derived fields}.
The derived physical fields can be calculated from the basic physical fields using the formulas provided in the table.
Figure \ref{derived_field_A} illustrates additional physical fields derived of case valid-A using the panoramic framework, all maintaining high accuracy.
\begin{table}[!ht]
  \renewcommand\arraystretch{1.2}
\centering
\caption{Explicit calculation method of various derived fields.}
\begin{tabular}{ccc}
\toprule
Fields name & Symbol & Expression formula  \\ \midrule
absolute total temperature & $t^*_{v}$ &  $t^*_{v}=t+\frac{v^2}{2C_p}$ \\
relative total temperature & $t^*_{w}$ &  $t^*_{w}=t+\frac{w^2}{2C_p}$ \\
absolute total pressure & $p^*_{v}$ &  $p^*_{v}=p(t^*_v/t)^{\kappa/\kappa-1}$ \\
relative total pressure & $p^*_{w}$ &  $p^*_{w}=p(t^*_w/t)^{\kappa/\kappa-1}$ \\
static entropy & ${H}$ & ${H}=\frac{Cp}{\kappa} \log \left(\left(\frac{p}{p_{\text {ref}}}\right)^{(1-\kappa)}\left(\frac{t}{t_{\text {ref}}}\right)^\kappa\right)$  \\
\bottomrule
\end{tabular}
\label{derived fields}
\end{table}

\subsubsection{Panoramic performance prediction}
Performance parameters typically reflect the overall performance of the turbomachinery and are key indicators in the design process.
In the panoramic framework, the required performance parameters can be calculated using the basic physical fields.
Table \ref{performance} lists nine commonly used performance metrics, all of which can be calculated using the basic physical fields selected in this study.
\begin{table}[htbp]
  \renewcommand\arraystretch{1.3}
\centering
\caption{Explicit calculation method of various performance metrics}
\begin{tabular}{ccc}
\toprule
Performance name & Symbol & Expression formula  \\ \midrule
isentropic efficiency & $\eta_{is}$ &  $\eta_{is}=
({\frac{p^*_{v2}}{p^*_{v1}}^{\frac{\kappa-1}{\kappa}}-1})/
({\frac{t^*_{v2}}{t^*_{v1}}-1})$ \\
polytropic efficiency & $\eta_{pol}$ &  $\eta_{pol}=\frac{R}{C_p} \cdot
{\ln \left(\frac{p^*_{v2}}{p^*_{v1}}\right)}/
{\ln \left(\frac{t^*_{t2}}{t^*_{v1}}\right)}$ \\
pressure ratio  & $\pi$ &  $\pi = p^*_{v2}/p^*_{v1}$ \\
temperature ratio  & $\theta$ &  $\theta = t^*_{v2}/t^*_{v1}$ \\
mass flow & $\dot{m}$ &  $\dot{m} = \rho v_m \cdot\pi(r_\text{s}^2 - r_\text{h}^2)$ \\
pressure loss coefficient & $\omega$ & ${\omega}=({p_{w2}^*-p_{w1}^*})/({p_{w1}^*-p_1})$  \\
isentropic temperature & $t^*_{is}$ & $t^*_{is}={t_{u1}^{*}}/{\left(\frac{p_{u1}}{p}\right)^{\frac{\kappa-1}{\kappa}}}$  \\
isentropic mach number & $\text{Ma}_{is}$ & $\text{Ma}_{is} = \sqrt{\left(\frac{p^*_{u1}}{p}^{(\kappa-1)/\kappa}-1  \right)\cdot\frac{2}{\kappa-1}}+\frac{u^2}{\kappa R_g t^*_{is}}$  \\
load coefficient & $\psi $ &  $\psi = C_p(t^*_{v1}-t^*_{v2})/u^2$ \\
\bottomrule
\end{tabular}
\label{performance}
\end{table}
\par
\begin{figure}[htbp]
\centering
\includegraphics[width=0.96\linewidth]{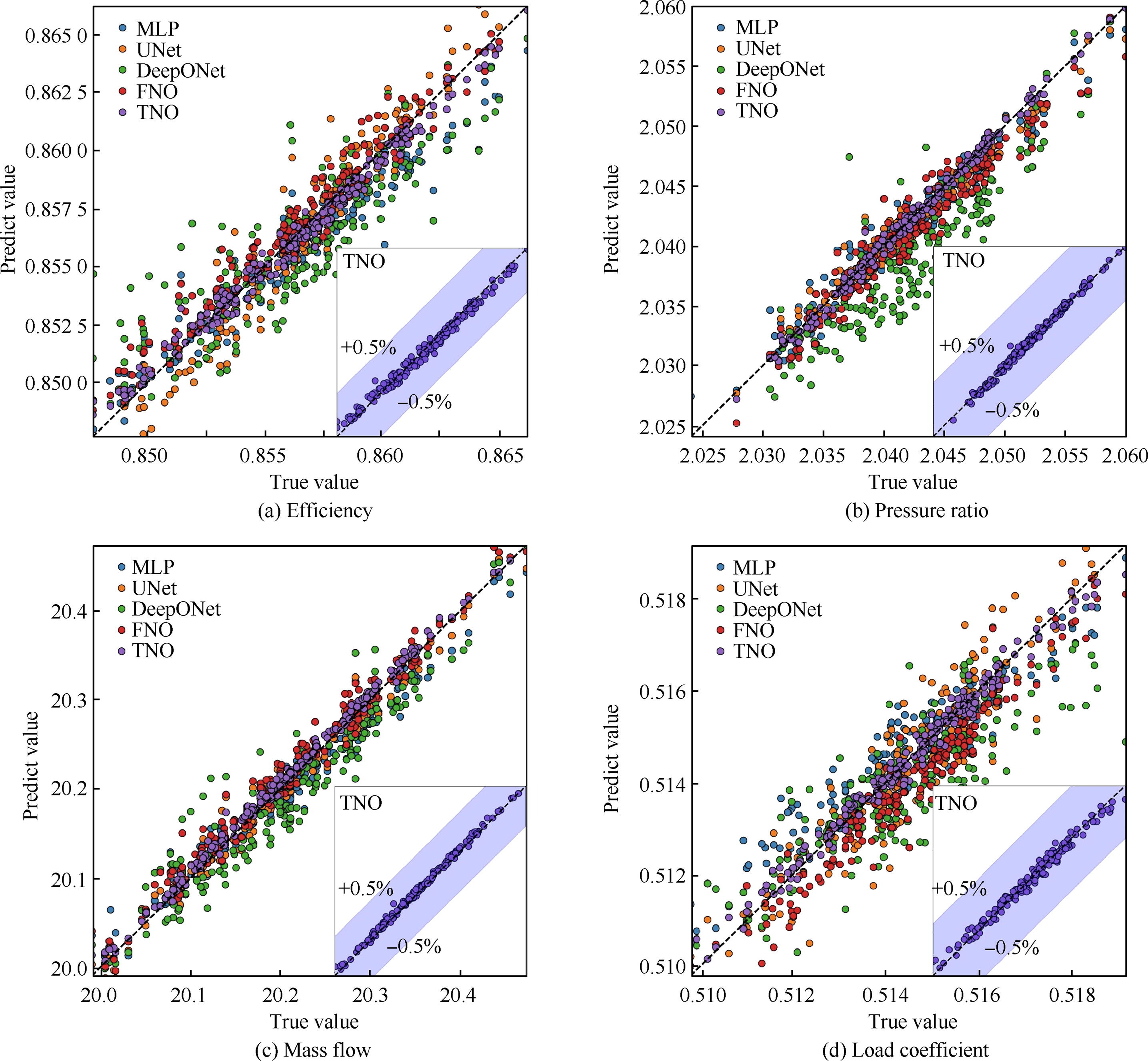}
\caption{Comparison between global performance values from CFD and the networks\textquotesingle{} predictions. The inset bands are $\pm0.5\%$. Reproduced from Fig. 11 of the published article under CC BY 4.0 (see title note).}
\label{globalperformance}
\end{figure}
\begin{figure}[htbp]
\centering
\includegraphics[width=0.96\linewidth]{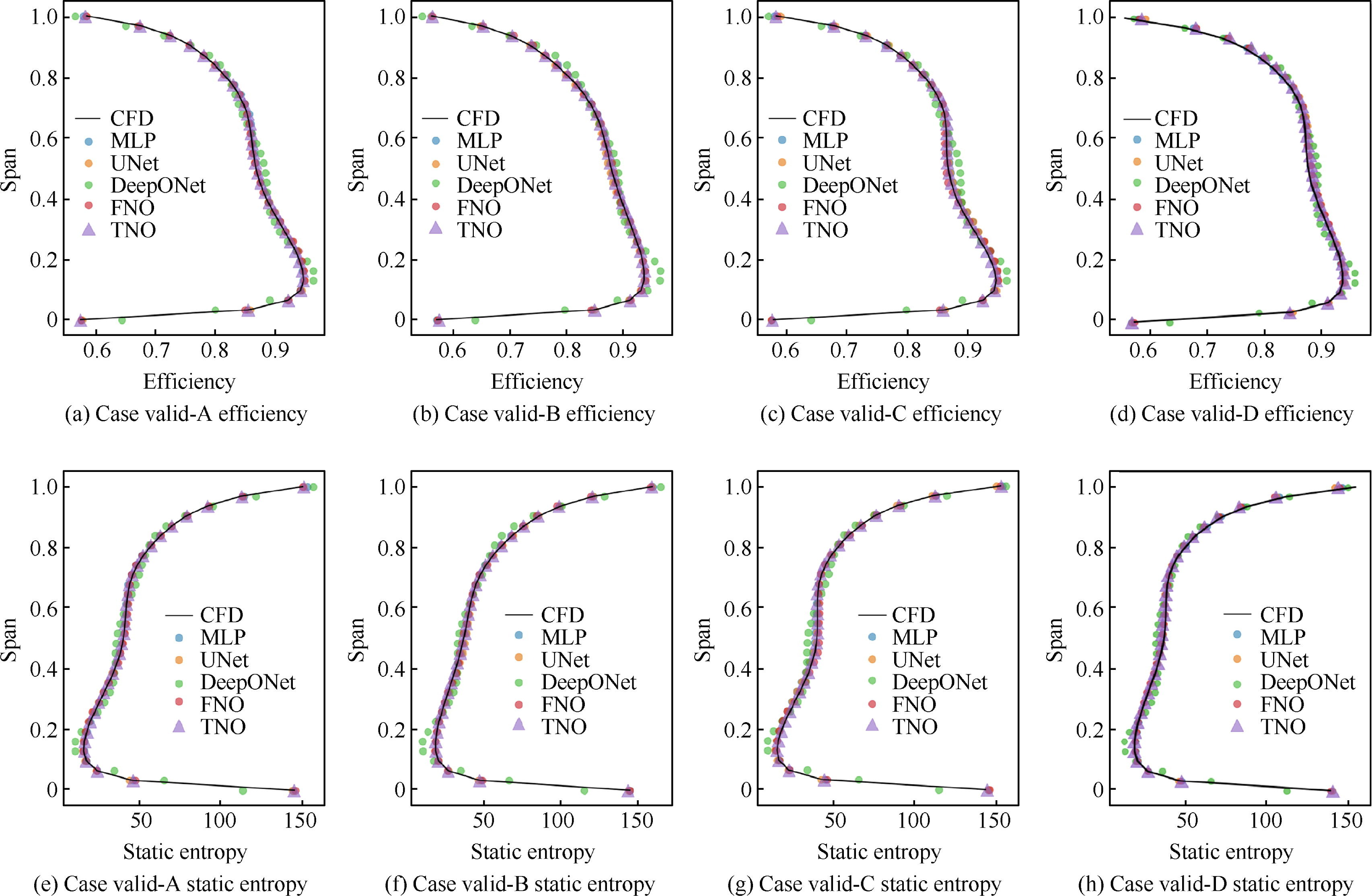}
\caption{Comparison between predicted and CFD spanwise performance curves for cases valid-A, valid-B, valid-C, and valid-D. Case D is the worst TNO prediction case selected from the validation set. Reproduced from Fig. 12 of the published article under CC BY 4.0 (see title note).}
\label{curves_compare}
\end{figure}

Figure \ref{globalperformance} illustrates the comparison between global performance values from the CFD and the network prediction for all validation samples.
Overall, the TNO network exhibits the best performance in predicting efficiency, pressure ratio and mass flow of the Rotor 37 blade, accurately predicting the performance of various designs in the validation set.
Figure \ref{curves_compare} presents the differences between the predicted and true distribution curves of performance metrics for cases valid-A, valid-B, valid-C, and valid-D.
The networks reproduce the overall spanwise trends. For the worst TNO case valid-D, efficiency is overpredicted around the middle of the span and static entropy is underpredicted, illustrating the remaining prediction error.
The aforementioned comparisons about scalars and curves of performance metrics lead to the conclusion that deep neural operator networks can accurately predict the meridional plane physical fields of a three-dimensional flow field.
\clearpage
\subsubsection{Comparison with conventional surrogates}
\par
\begin{figure}[!ht]
\begin{center}
\includegraphics[scale=0.25, trim = 3.6cm 1cm 3.6cm 2cm, clip]{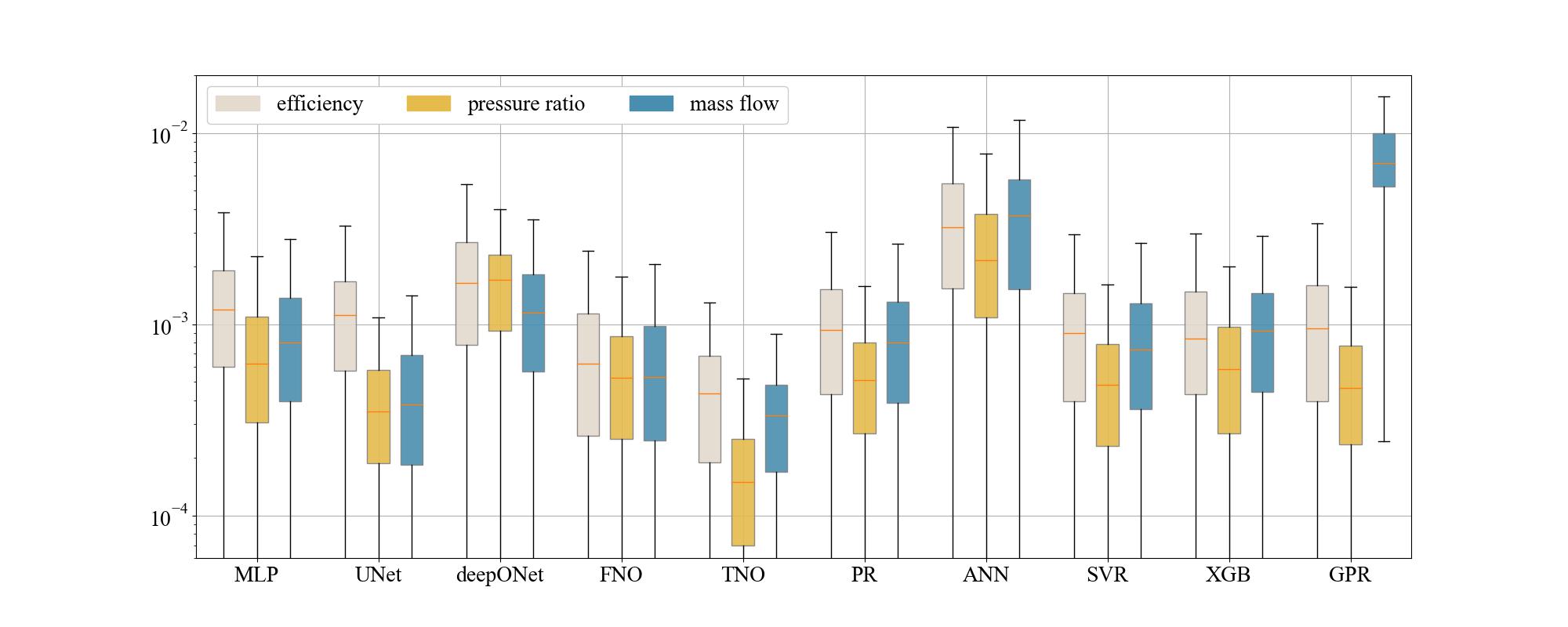}
\end{center}
\caption{The comparison between the accuracy of deep neural networks and conventional surrogate models}
\label{compare_surrogate}
\end{figure}

In this section, 5 conventional surrogate modeling methods, and were compared with above deep neural networks.
The surrogates used in the comparison include: ANN, PR, GPR, SVR, and XGB, optimized hyperparameter settings of 5 surrogates are shown in Table \ref{surrogate_set}.
\begin{table}[!ht]
\centering
\small
\caption{Conventional surrogate model hyperparameters and implementation libraries}
\setlength{\tabcolsep}{3pt}
\begin{tabular}{p{0.075\linewidth}>{\raggedright\arraybackslash}p{0.51\linewidth}>{\raggedright\arraybackslash}p{0.34\linewidth}}
\toprule
Model & Hyperparameter setting & Library \\ \midrule
PR & degree=3 & sklearn.linear\_model.\newline LinearRegression \\ \midrule
ANN & depth=5; width=32; learning rate=0.01; iter\_max=500 & torch.nn.Module \\ \midrule
SVR & kernel:linear; degree=3; C=1.0; epsilon=0.1; cache\_size=200 & sklearn.svm.SVR \\ \midrule
XGB & objective:reg:squarederror; max\_depth=6; n\_estimators=160; subsample=0.8; reg\_lambda=1e-5; learning rate=0.1 & xgboost.XGBRegressor \\ \midrule
GPR & kernel:scale-RBF; likelihood:gaussianLikelihood; variational distribution:multivariateNormal; learning rate=0.1; iter\_max=500 & gpytorch.models.\newline ExactGP \\ \bottomrule
\end{tabular}
\label{surrogate_set}
\end{table}
The definition of the performance related error (PRE) is $e_\text{PRE} = \sum_{i=1}^{N_\text{valid}}{\sqrt{(\mathbf{\phi}_{i}-\hat{\mathbf{\phi}}_{i})^2}}/{|\mathbf{\phi}_{i}|}/{N_\text{valid}}$.
Figure \ref{compare_surrogate} illustrates the comparison of prediction accuracy between surrogate models and the deep neural networks.
Each Network employs single model to predict the basic flow fields and subsequently calculate the performance metrics, whereas each surrogate model directly predict the three performance metrics independently.
As observed in the figure, the deep neural networks outperform the conventional surrogates in predicting all kinds of three performance metrics.
\par
Furthermore, the relationship between training sample size and prediction accuracy was investigated by conducting training with varying-sized training sets for both the 5 deep neural networks and the 5 surrogate models.
The prediction errors in the validation set are presented in Table \ref{compare_surrogate_data}, with the optimal prediction results for each scenario highlighted in bold.
The data clearly demonstrate the advantages of the deep neural operator model.
Across the 15 metric--sample-size scenarios in Table \ref{compare_surrogate_data}, TNO has the lowest prediction error in 12 scenarios, including every scenario with at least 1500 training samples.
\begin{table}[htbp]
\centering
\caption{The predict accuracy of deep neural networks and conventional surrogate models with different training sample size}
\resizebox{\linewidth}{!}{
\begin{tabular}{ccccccc}
\toprule
performanca & Algo & 500 & 1000 & 1500 & 2000 & 2500  \\  \midrule
efficiency & MLP & 1.064\textperthousand & 1.091\textperthousand & 1.042\textperthousand & 1.031\textperthousand & 1.318\textperthousand  \\
  ~ & UNet & 1.350\textperthousand & 1.980\textperthousand & 1.331\textperthousand & 1.245\textperthousand & 1.253\textperthousand  \\
  ~ & deepONet & 163.1\textperthousand & 88.92\textperthousand & 27.56\textperthousand & 6.535\textperthousand & 1.840\textperthousand  \\
  ~ & FNO & 1.213\textperthousand & 1.188\textperthousand & 1.057\textperthousand & 0.938\textperthousand & 0.810\textperthousand  \\
  ~ & \bf{TNO} & \bf{0.823}\textperthousand & \bf{0.460}\textperthousand & \bf{0.448}\textperthousand & \bf{0.439}\textperthousand & \bf{0.486}\textperthousand  \\
  ~ & POLY & 1.086\textperthousand & 1.079\textperthousand & 1.048\textperthousand & 1.070\textperthousand & 1.069\textperthousand  \\
  ~ & ANN & 3.697\textperthousand & 3.775\textperthousand & 3.799\textperthousand & 3.699\textperthousand & 3.706\textperthousand  \\
  ~ & SVR & 1.040\textperthousand & 1.029\textperthousand & 1.035\textperthousand & 1.031\textperthousand & 1.03\textperthousand  \\
  ~ & XGB & 1.666\textperthousand & 1.370\textperthousand & 1.237\textperthousand & 1.123\textperthousand & 1.032\textperthousand  \\
  ~ & GPR & 1.425\textperthousand & 1.273\textperthousand & 1.156\textperthousand & 1.098\textperthousand & 1.075\textperthousand  \\
  \midrule
  pressure ratio & MLP & 0.803\textperthousand & 0.926\textperthousand & 0.929\textperthousand & 0.989\textperthousand & 0.768\textperthousand  \\
  ~ & UNet & 0.854\textperthousand & 0.656\textperthousand & 0.955\textperthousand & 0.556\textperthousand & 0.413\textperthousand  \\
  ~ & deepONet & 134.7\textperthousand & 125.4\textperthousand & 37.30\textperthousand & 0.910\textperthousand & 1.638\textperthousand  \\
  ~ & FNO & 1.668\textperthousand & 1.511\textperthousand & 0.932\textperthousand & 1.794\textperthousand & 0.600\textperthousand  \\
  ~ & \bf{TNO} & 1.089\textperthousand & 1.024\textperthousand & \bf{0.513}\textperthousand & \bf{0.509}\textperthousand & \bf{0.185}\textperthousand  \\
  ~ & POLY & 0.596\textperthousand & 0.597\textperthousand & 0.592\textperthousand & 0.598\textperthousand & 0.595\textperthousand  \\
  ~ & ANN & 3.026\textperthousand & 2.477\textperthousand & 2.802\textperthousand & 2.750\textperthousand & 2.593\textperthousand  \\
  ~ & SVR & \bf{0.579}\textperthousand & \bf{0.579}\textperthousand & 0.581\textperthousand & 0.582\textperthousand & 0.576\textperthousand  \\
  ~ & XGB & 1.134\textperthousand & 0.901\textperthousand & 0.788\textperthousand & 0.726\textperthousand & 0.684\textperthousand  \\
  ~ & GPR & 0.651\textperthousand & 0.621\textperthousand & 0.587\textperthousand & 0.573\textperthousand & 0.556\textperthousand  \\
  \midrule
  mass flow & MLP & 1.065\textperthousand & 0.948\textperthousand & 0.926\textperthousand & 0.862\textperthousand & 0.961\textperthousand  \\
  ~ & UNet & \bf{0.655}\textperthousand & 0.639\textperthousand & 0.621\textperthousand & 0.525\textperthousand & 0.493\textperthousand  \\
  ~ & deepONet & 224.8\textperthousand & 263.7\textperthousand & 50.97\textperthousand & 2.885\textperthousand & 1.235\textperthousand  \\
  ~ & FNO & 1.336\textperthousand & 1.054\textperthousand & 0.636\textperthousand & 0.614\textperthousand & 0.682\textperthousand  \\
  ~ & \bf{TNO} & 0.889\textperthousand & \bf{0.434}\textperthousand & \bf{0.495}\textperthousand & \bf{0.266}\textperthousand & \bf{0.361}\textperthousand  \\
  ~ & POLY & 0.949\textperthousand & 0.926\textperthousand & 0.932\textperthousand & 0.931\textperthousand & 0.930\textperthousand  \\
  ~ & ANN & 3.929\textperthousand & 4.029\textperthousand & 4.215\textperthousand & 3.962\textperthousand & 4.024\textperthousand  \\
  ~ & SVR & 0.910\textperthousand & 0.895\textperthousand & 0.908\textperthousand & 0.904\textperthousand & 0.913\textperthousand  \\
  ~ & XGB & 1.688\textperthousand & 1.389\textperthousand & 1.226\textperthousand & 1.145\textperthousand & 1.015\textperthousand  \\
  ~ & GPR & 7.269\textperthousand & 7.273\textperthousand & 7.284\textperthousand & 8.37\textperthousand & 7.624\textperthousand  \\
\bottomrule
\end{tabular}}
\label{compare_surrogate_data}
\end{table}
In the following content, the TNO model with the best overall performance using 2500 samples will be applied to various downstream tasks to further test its effect.
\subsection{Performance sensitivity analysis of the compressor}
Sensitivity analysis is widely used in aerodynamic design, and often relies on surrogate models to accomplish the task~\cite{shahsavaniVariancebasedSensitivityAnalysis2011}.
So, based on the comprehensive flow field prediction method in this paper, it is possible to further analyze the relationship between design variables and performance metrics variations compared with conventional surrogate models.
\par
\begin{figure}[!ht]
\begin{center}
\includegraphics[scale=0.28, trim = 10cm 2.5cm 10cm 2.5cm]{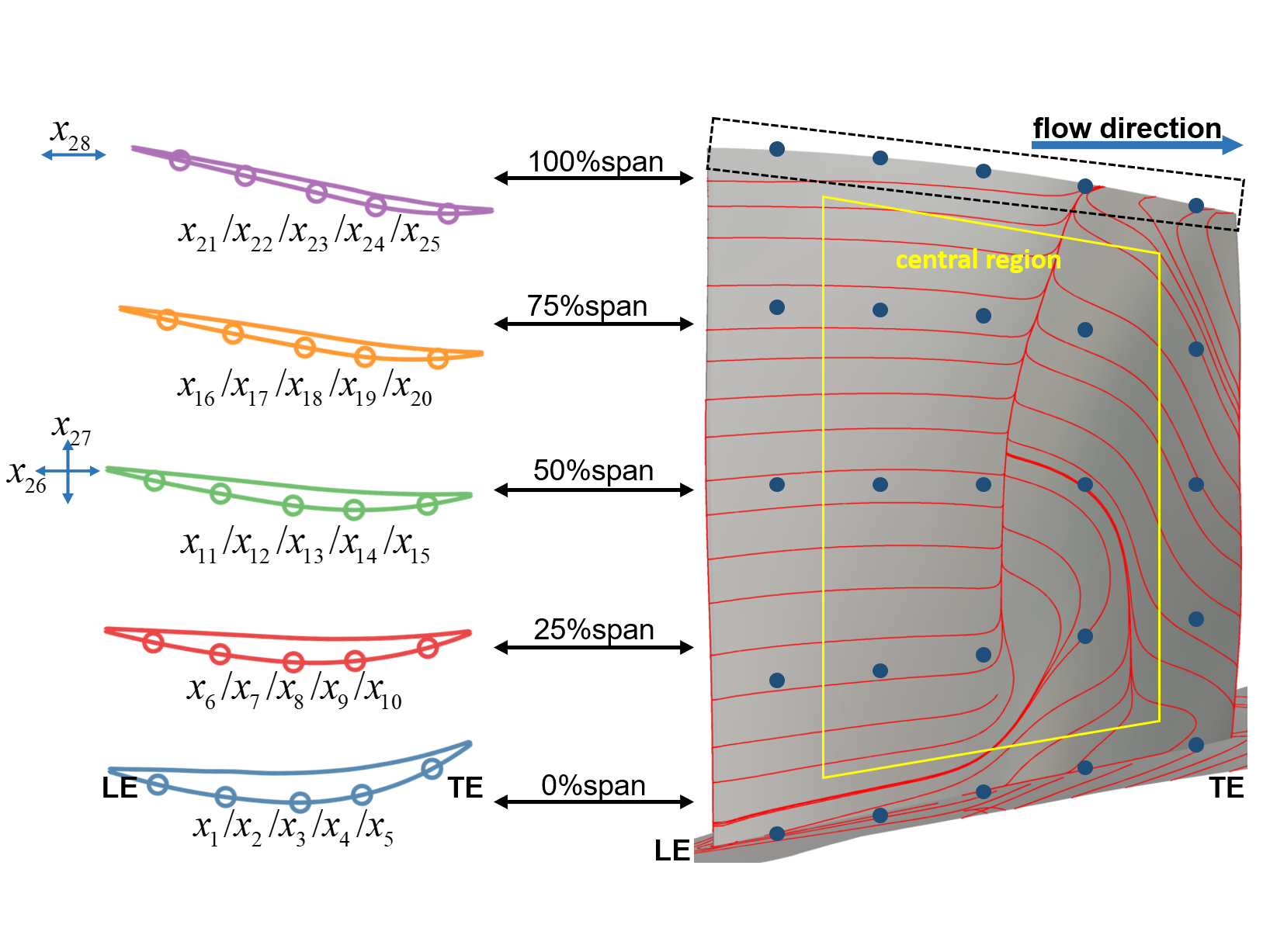}
\end{center}
\caption{The influence area of each design variable on the blade suction surface.}
\label{var_loc}
\end{figure}

This paper employs the variance-based sensitivity analysis technique using the Monte Carlo method~\cite{borgonovoSensitivityAnalysisReview2016} to obtain the local sensitivity.
Fig. \ref{var_loc} illustrates the influence area of each variable on the blade suction surface.
the 28 design variables are grouped based on blade height in following analysis.
\par
\begin{figure}[htbp]
\centering
\includegraphics[width=0.49\linewidth]{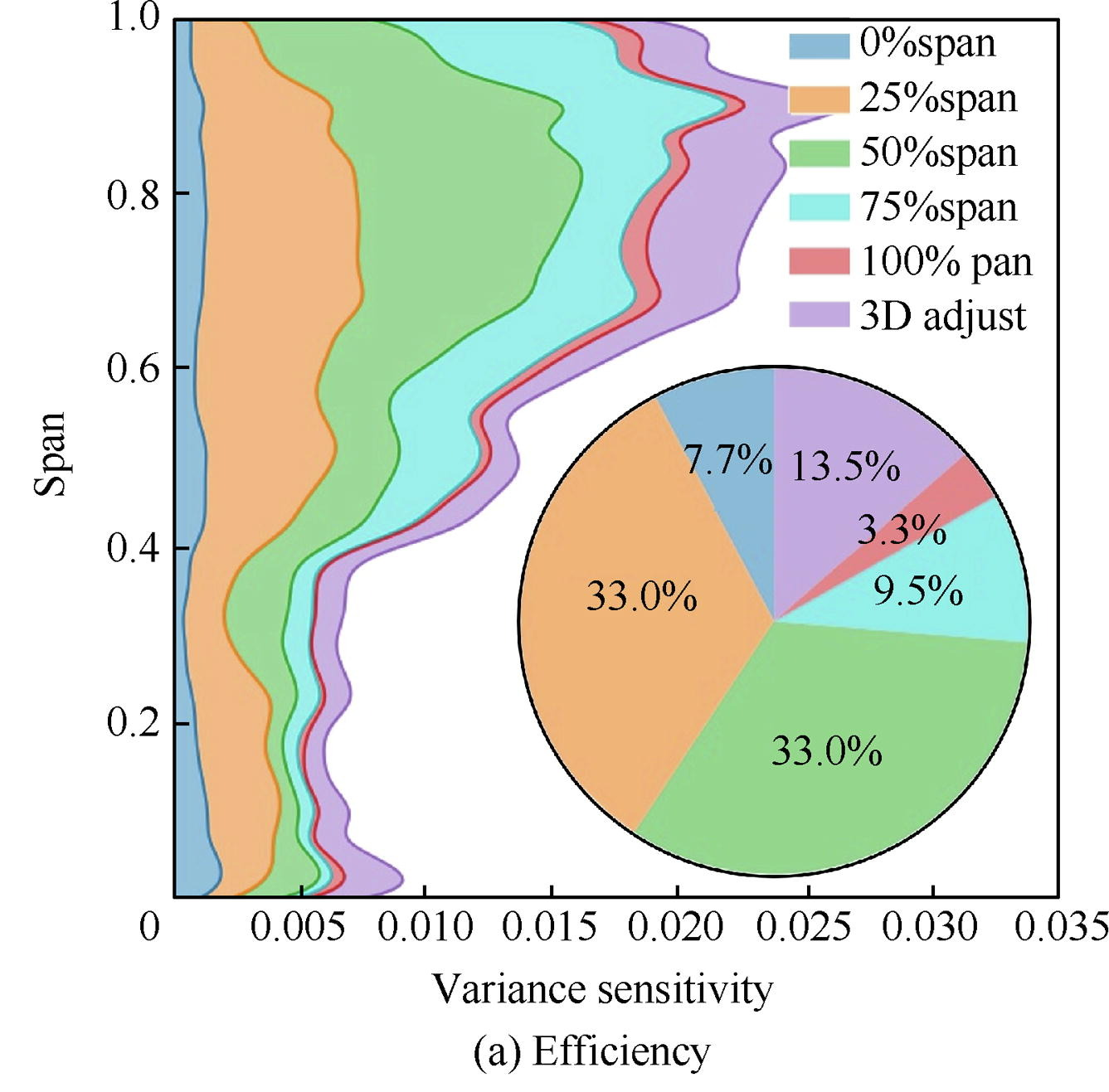}\hfill
\includegraphics[width=0.49\linewidth]{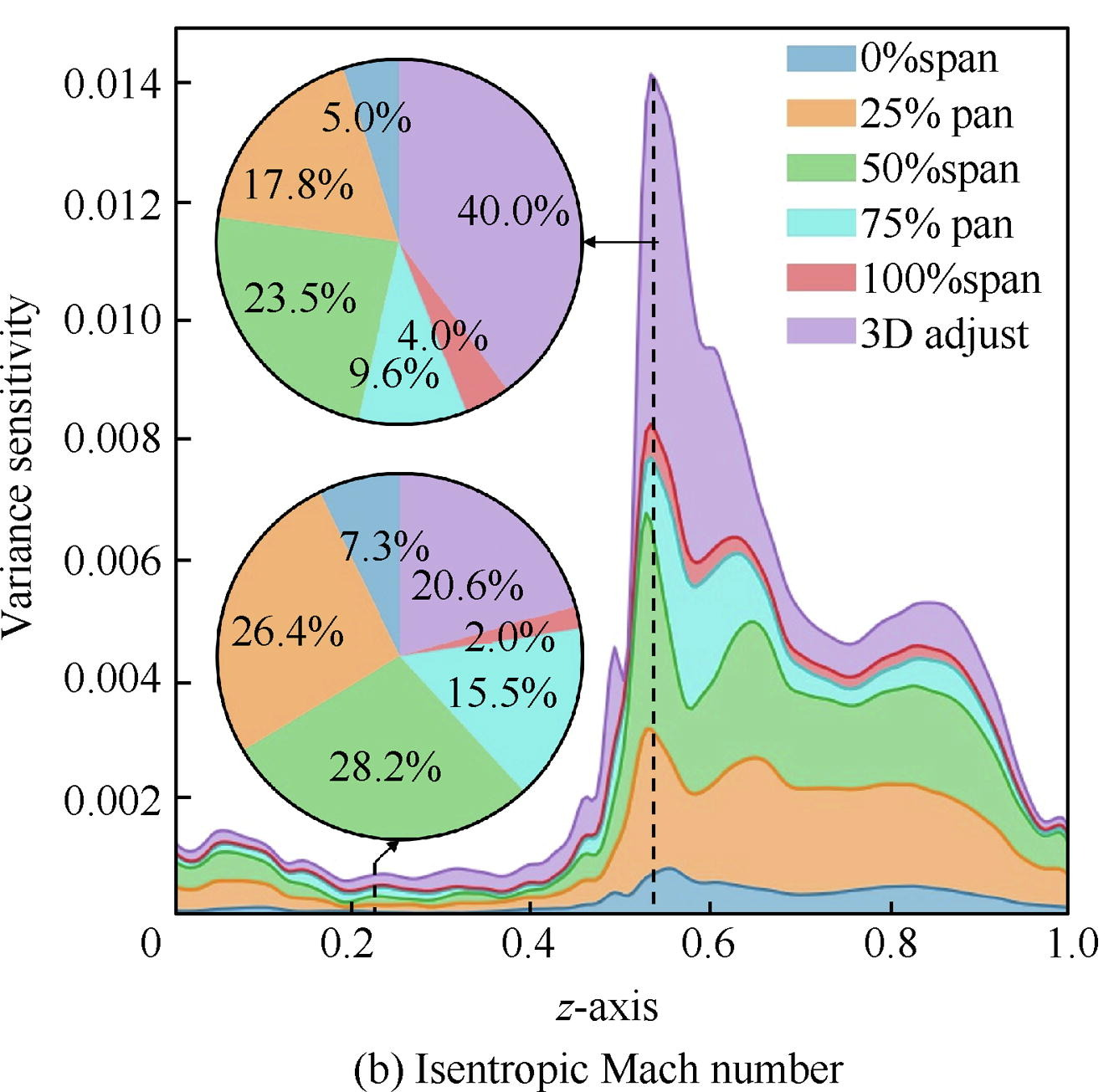}
\caption{Sensitivity analysis of the efficiency span distribution and isentropic Mach number flow distribution. Reproduced from Fig. 14 of the published article under CC BY 4.0; the two original panels are arranged side by side (see title note).}
\label{height sensitive}
\end{figure}

Fig. \ref{height sensitive}(a) shows sensitivity analysis result of the Rotor 37 blade's efficiency span distribution.
Overall, the variables have a more pronounced impact on the upper flow region of the passage.
As shown in the pie chart, variable groups span 25\% and span 50\% exhibit similar overall effects on the overall efficiency of Rotor 37, .
However, a further analysis using the panoramic flow field prediction model reveals that these two groups actually have completely different forms of influence on the flow field.
Group span 25\% mainly affects the $5\% \sim 25\%$ and $50\%\sim 90\%$ regions, where the former represents the actual impact of blade geometry at the same blade height position, and the latter represents the manifestation of local profile changes that affect the flow and spread globally through the separation vortex.
\par
Figure \ref{height sensitive}(b) shows the variance influence accumulation diagram of the flow direction distribution of isentropic Mach number.
In terms of the influence of different variables on isentropic Mach number, it is found that in the region with different flow direction, the influence ratio of different groups of variables on isentropic Mach number is obviously different.
\par
With more detailed analysis, designers can gain a more accurate understanding of the range of influence and interactions between different variable groups, enabling them to make more precise and targeted design adjustments.
\clearpage
\subsection{ Fast and flexible scenario optimization}
\par
In this section, the single-objective and the multi-objective optimization design tasks are carried out using the trained TNO network.
The optimization design employed the genetic algorithm (GA)~\citep{mitchellIntroductionGeneticAlgorithms1998}, while setting the population size as 20 and the generation number as 100.
The results of 5 single-objective optimization tasks are presented in Table \ref{single_opt}.
All obtained optimal results underwent verification using CFD simulations.
\begin{table}[htbp]
\centering
\caption{The optimization and verify results of single-objective optimization tasks.}
\resizebox{\linewidth}{!}{
\begin{tabular}{cccccccc}
\toprule
\makecell[c]{performance \\parameter} & \makecell[c]{search\\ objective} & \makecell[c]{reference\\ value} & \makecell[c]{optimal \\predict value} & \makecell[c]{optimal \\CFD value}  & \makecell[c]{related\\ error} &  \makecell[c]{related\\ improvement} \\
\midrule
efficiency & max & 0.854  & 0.871 & 0.871 & -0.001\% & 1.95\%  \\
pressure loss  & min & 0.124 & 0.115 & 0.103 & 11.87\% & 17.24\%  \\
static entropy & min & 44.83 & 41.44 & 38.94 & 6.42\% & 13.13\%  \\
pressure ratio & max & 2.046 & 2.073 & 2.071 & 0.11\% & 1.21\%  \\
temperature ratio & max & 1.265 & 1.270 & 1.271 & -0.03\% & 0.44\%  \\
\bottomrule
\end{tabular}}
\label{single_opt}
\end{table}
\par
\begin{figure}[htbp]
\centering
\subfigure[reference]{
\begin{minipage}[t]{0.22\linewidth}
\centering
\includegraphics[width=1\textwidth]{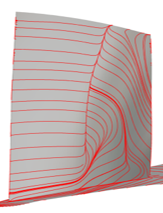}
\end{minipage}%
}%
\subfigure[opt-A]{
\begin{minipage}[t]{0.22\linewidth}
\centering
\includegraphics[width=1\textwidth]{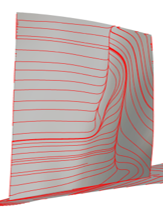}
\end{minipage}
}%
\subfigure[opt-B]{
\begin{minipage}[t]{0.22\linewidth}
\centering
\includegraphics[width=1\textwidth]{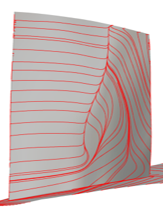}
\end{minipage}%
}%
\subfigure[opt-C]{
\begin{minipage}[t]{0.22\linewidth}
\centering
\includegraphics[width=1\textwidth]{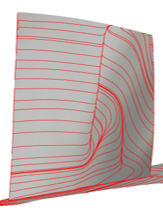}
\end{minipage}
}%

\centering
\caption{The suction surface limit flow diagrams of the reference and optimal samples}
\label{streamlines}
\end{figure}

In comparison to our prior work on the similar optimization task~\cite{wangKTEGOKnowledgeTransfer2022}, the obtained optimal efficiency values exhibit a general consistency, as the actual improvement of efficiency is 1.70\%, validating the effectiveness of the panoramic prediction method.
The panoramic prediction framework successfully achieves an optimization task that previously necessitated 1000 sets of CFD calculations, now accomplished within a mere 121 seconds, thereby enhancing the optimization efficiency by four orders of magnitude.
Multi-objective optimization is inherently more challenging than the single-objective.
Three multi-objective optimization tasks are carried to verified the trained TNO network in this section.
The NSGA2~\cite{debFastElitistMultiobjective2002} algorithm is used in these tasks, employing a population size of 30 and a generation number of 100.
The results of 3 multi-objective optimization tasks are presented in Table \ref{optimalRst}.
All obtained optimal results underwent verification using CFD simulation.
For the solution sets of the above three optimization tasks, the samples with the highest efficiency were selected for further analysis, which were denoted as opt-A, opt-B, and opt-C respectively.
Table \ref{optimalRst} shows the reference design and the different performance metrics for the three samples.
On one hand, these performance indicators demonstrate the diversity and accuracy achievable through panoramic forecasting methods.
On the other hand, a comparison with the reference design reveals three distinct characteristics of the optimization results: efficiency improved in all three optimal designs (including isentropic and polytropic efficiency) and reduced losses (including total pressure loss and entropy).
\begin{table}[htbp]
  \centering
  \caption{Performance metrics of reference and optimal designs}
  \resizebox{\linewidth}{!}{
  \begin{tabular}{ccccc}
    \toprule
    \tabincell{c}{setting and \\ performance} & ref & opt-A & opt-B & opt-C  \\
    \midrule
    objective & - & $\max\eta_{is},\max \pi$  & $\max\eta_{is},\min\pi$ & $\max\eta_{is},\min\pi$\\
    constrain & - & -  & - & $\dot{m} \leq 20.265 $\\
    \midrule
    isentropic efficiency  & 0.854(-0.98\textperthousand) & 0.871(-0.37\textperthousand) & 0.869(1.66\textperthousand) & 0.863(-1.76\textperthousand)  \\
    polytropic efficiency & 0.872(-0.89\textperthousand) & 0.883(-0.30\textperthousand) & 0.882(1.48\textperthousand) & 0.876(-1.58\textperthousand)  \\
    pressure loss  & 0.124(3.56\textperthousand) & 0.116(7.265\textperthousand) & 0.116(5.88\textperthousand) & 0.119(14.7\textperthousand)  \\
    pressure ratio & 2.046(-1.04\textperthousand) & 2.069(-0.93\textperthousand) & 2.045(-0.68\textperthousand) & 2.04(-0.66\textperthousand)  \\
    temperature ratio & 1.265(-0.13\textperthousand) & 1.266(-0.30\textperthousand) & 1.262(-0.64\textperthousand) & 1.263(0.15\textperthousand)  \\
    mass flow & 20.266(-0.64\textperthousand) & 20.615(0.42\textperthousand) & 20.439(0.66\textperthousand) & 20.262(-0.45\textperthousand)  \\
    entropy & 44.832(5.65\textperthousand) & 41.196(-5.08\textperthousand) & 41.273(-15.1\textperthousand) & 43.462(9.48\textperthousand)  \\
    max mach number & 1.712(0.12\textperthousand) & 1.722(-0.02\textperthousand) & 1.717(0.21\textperthousand) & 1.71(0.07\textperthousand)  \\
    load coefficient & 0.514(-0.55\textperthousand) & 0.515(-2.03\textperthousand) & 0.507(-2.95\textperthousand) & 0.51(0.73\textperthousand)  \\
  \bottomrule
  \end{tabular}}
  \label{optimalRst}
  \end{table}
Figure. \ref{streamlines} shows the reference design and the suction surface limit flow diagrams of the three samples respectively.
In conjunction with Fig. \ref{streamlines}, it is evident that the area of the suction surface separation region exhibits varying degrees of reduction after optimization, aligning with the performance improvements.
The comparison between pressure ratio and mass flow rate confirmed a strong positive correlation between these two variables in the design of Rotor 37 blade, consistent with the literature ~\cite{guoResearchMetaModelBased2015}.
Opt-A and opt-B exhibit significantly higher mass flow rates compared to the reference design, whereas opt-C maintains a mass flow rate controlled at a level consistent with the reference design through imposed constraints.
\par
\begin{figure}[htbp]
  \centering

  \subfigure[efficiency-pressure ratio]{
  \begin{minipage}[t]{0.5\linewidth}
  \centering
  \includegraphics[width=1\textwidth]{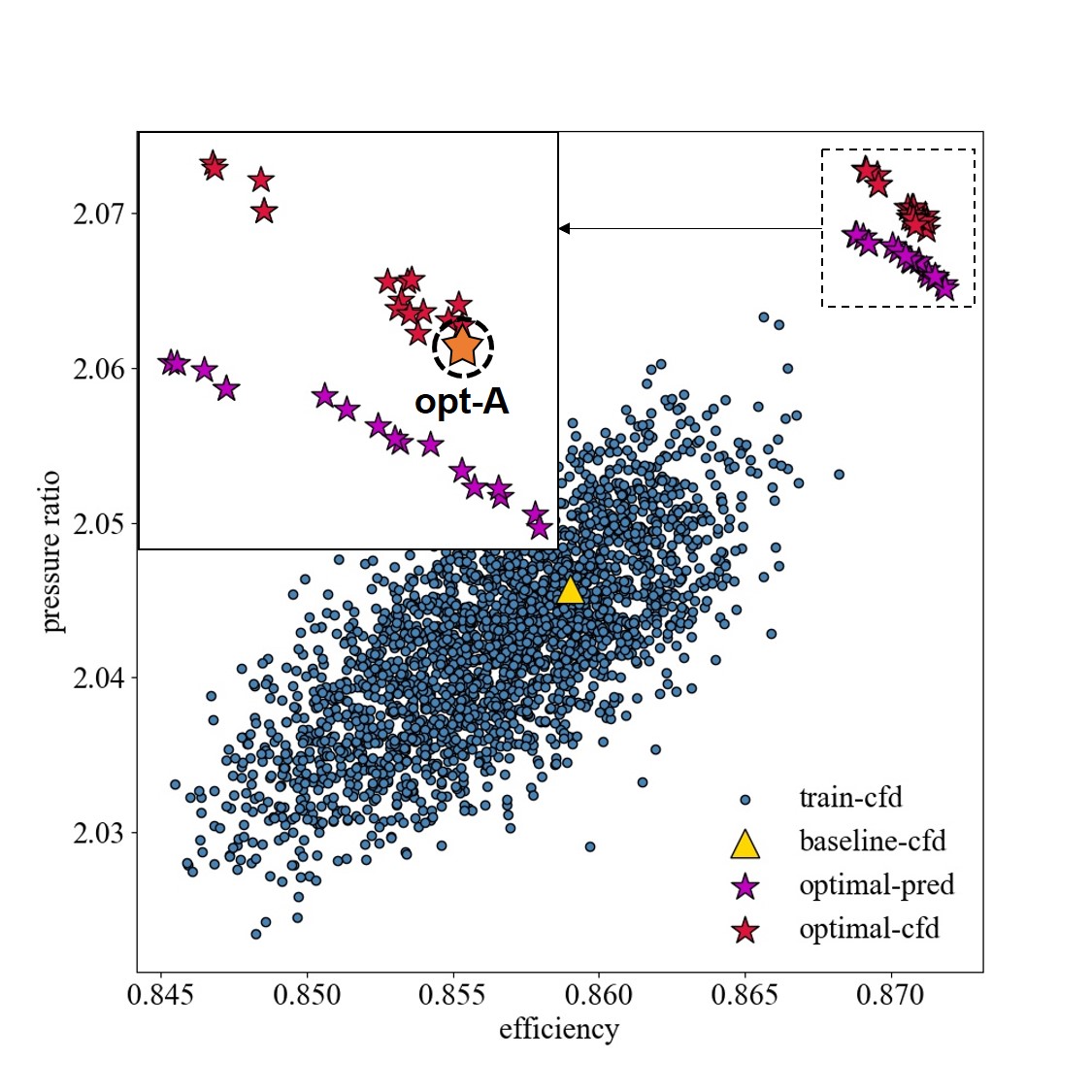}
  \end{minipage}%
  }%
  \subfigure[efficiency-mass flow]{
  \begin{minipage}[t]{0.5\linewidth}
  \centering
  \includegraphics[width=1\textwidth]{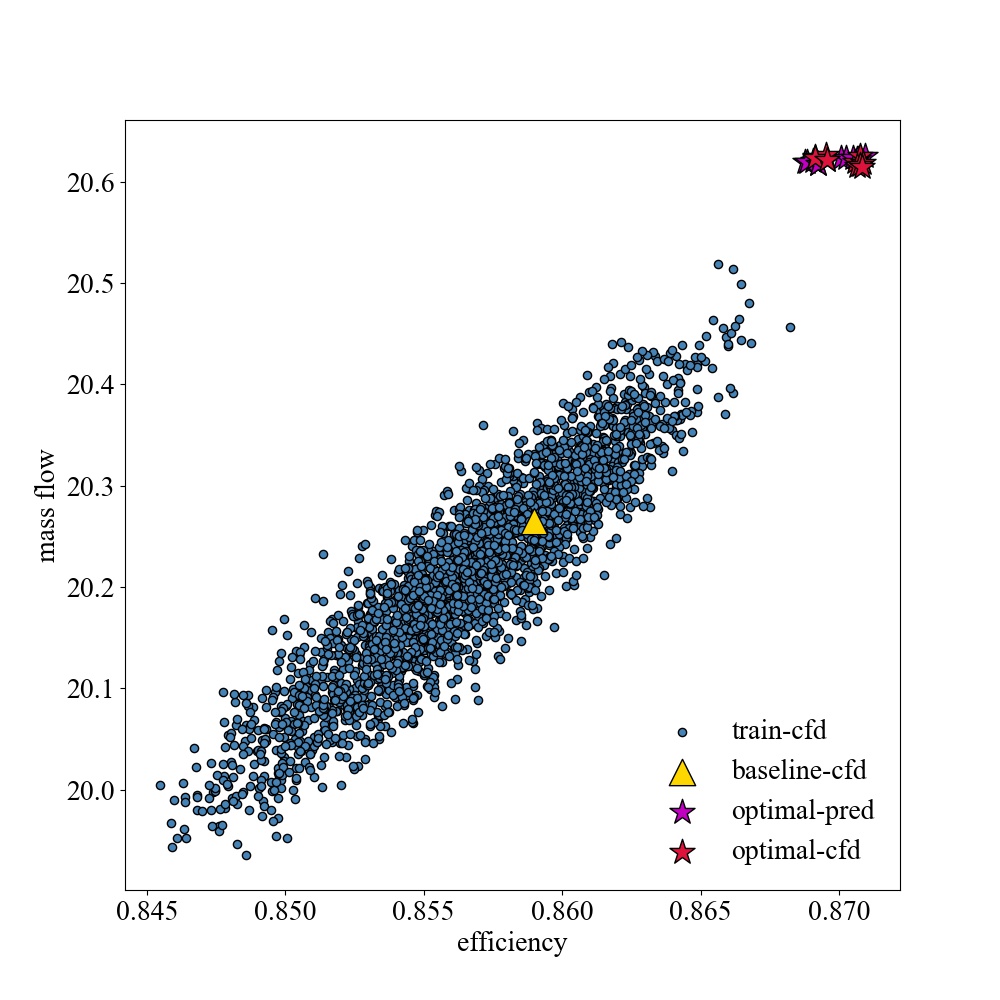}
  \end{minipage}
  }%

  \centering
  \caption{Optimal results of multi-objective optimization task-1}
  \label{optimal_task_1}
  \end{figure}

Figure \ref{optimal_task_1} illustrates the outcomes achieved for the multi-objective optimization task-1,
where the blue dots denote the distribution of the entire training set, while the yellow triangle denote the performance of the reference design.
The red stars correspond to the predicted Pareto solution set, which were further verified by CFD simulations denoted as purple stars.
The figure clearly demonstrates that the optimization process successfully identified a design surpassing the reference design in terms of both efficiency and pressure ratio.
A comparison between the training samples and the optimal solution reveals that the performance value of the identified optimal solution significantly exceeds the distribution range of the corresponding performance indicators observed in the training sample.
\par
\begin{figure}[htbp]
\centering
\subfigure[task-2 efficiency-pressure ratio]{
\begin{minipage}[t]{0.5\linewidth}
\centering
\includegraphics[width=1\textwidth]{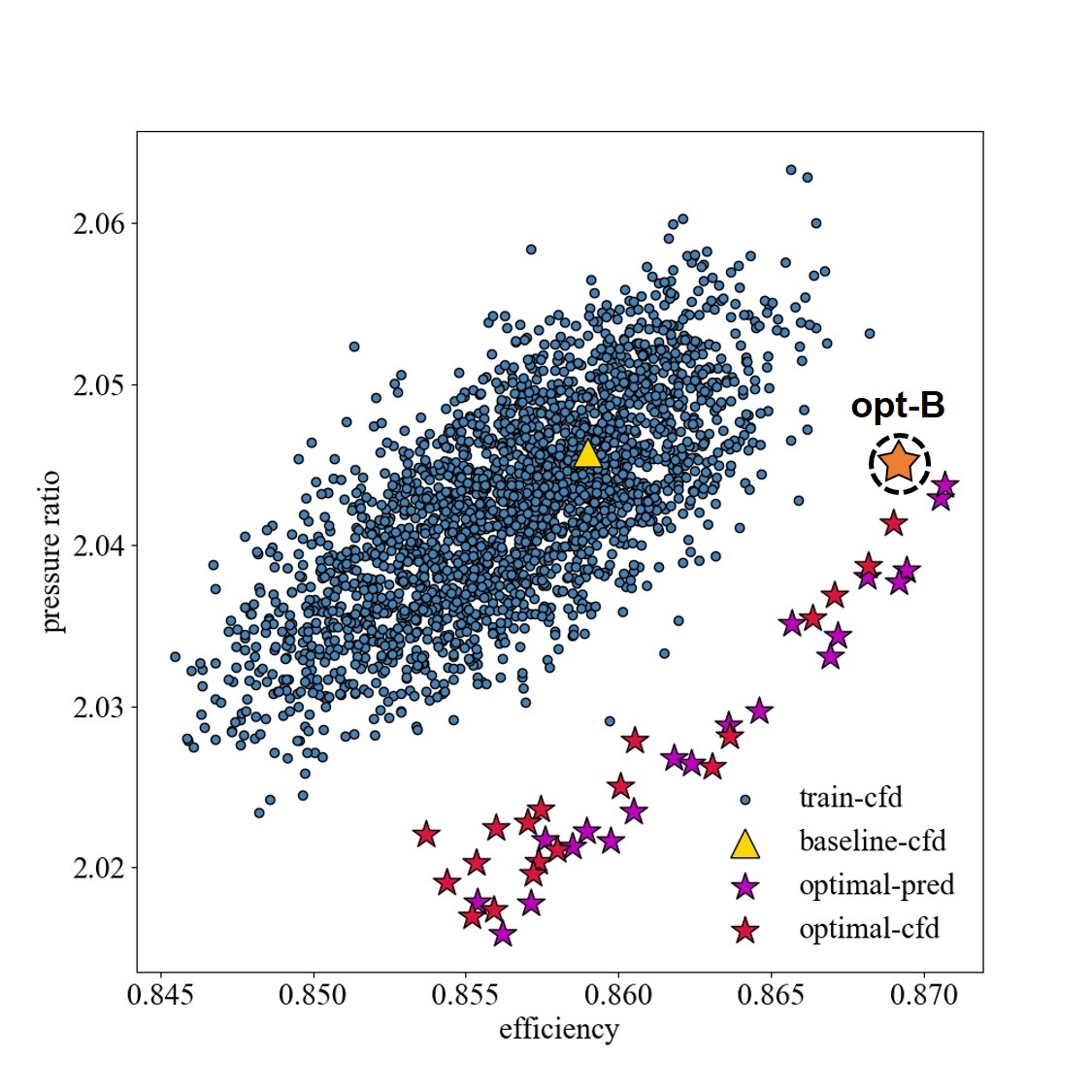}
\end{minipage}%
}%
\subfigure[task-3 efficiency-pressure ratio]{
\begin{minipage}[t]{0.5\linewidth}
\centering
\includegraphics[width=1\textwidth]{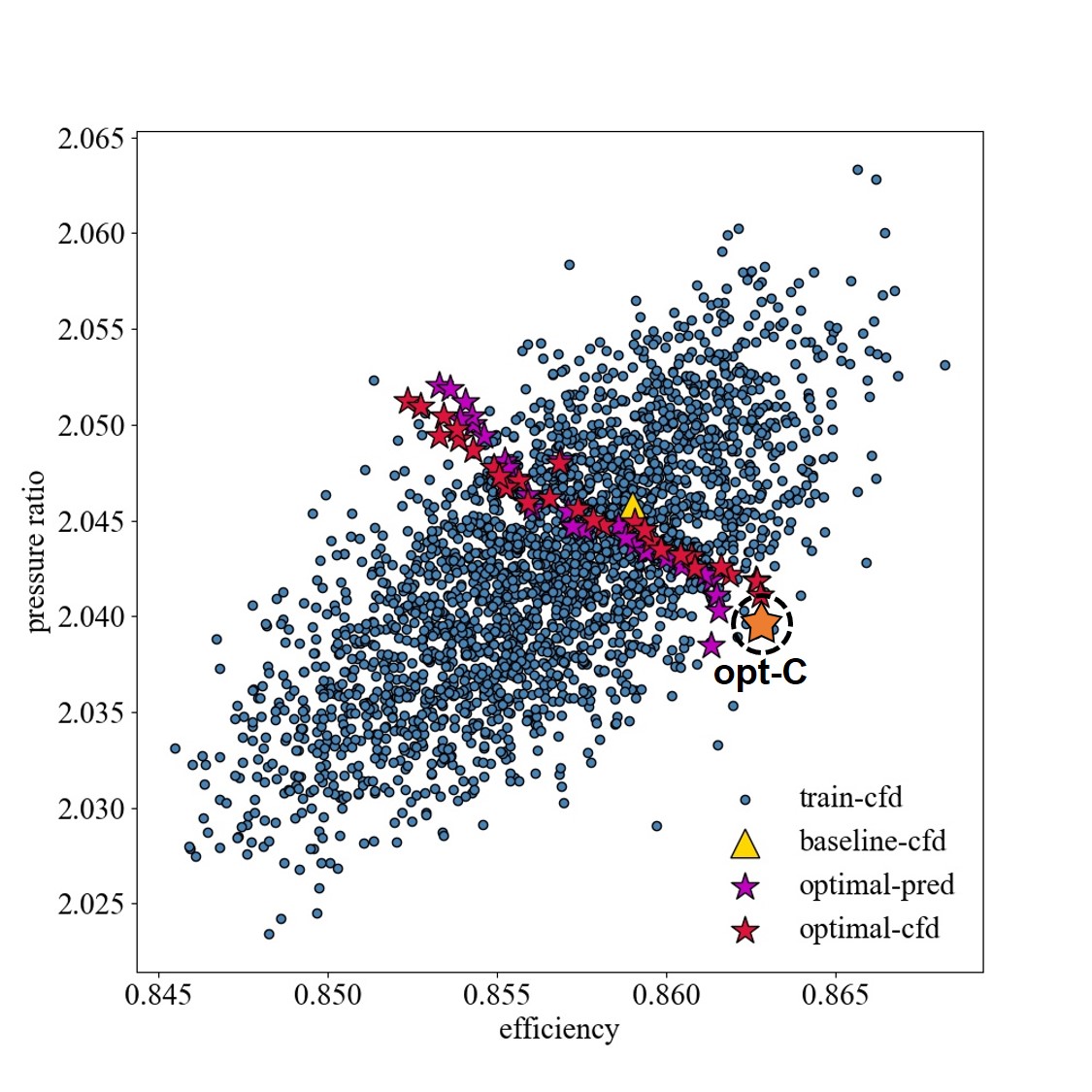}
\end{minipage}
}%

\centering
\caption{Optimal results of multi-objective optimization task-2 and task-3}
\label{optimal_task_2}
\end{figure}

Figure \ref{optimal_task_2} depicts the optimization results for task-2 and 3.
The successful completion of the aforementioned optimization tasks demonstrates the flexibility of the panoramic prediction method in facilitating optimization design.
\par
Base on the above content about the design process of Rotor 37 blade, the effectiveness of the panoramic prediction framework in turbomachinery design has been convincingly demonstrated.
Utilizing a panoramic framework, various downstream tasks can be accomplished with great efficiency.
Moreover, the model exemplifies its rapidity and flexibility by allowing for the arbitrary establishment of optimization objectives and constraints, thereby expediting the attainment of optimal solutions.
\clearpage
\section{Conclusions}
\par
The aerodynamic design of turbomachinery is a key challenge in aerospace engineering.
To expedite various downstream tasks within the aerodynamic design cycle, this paper introduces a panoramic prediction framework for turbomachinery.
Central to this framework is a novel Transformer-enhanced neural operator network, termed TNO.
Within the panoramic framework, TNO predicts the basic physical field governed by the Navier-Stokes equations, enabling the panoramic predictor to derive and compute any desired performance parameters.
This design achieves the goal of reusing a network across multiple downstream tasks, significantly improving efficiency.
\par
The proposed panoramic framework was tested on a typical transonic compressor, Rotor 37 blade.
The inference speed of the panoramic flow prediction model, based on the neural operator network, is much faster than that of the CFD numerical method.
Compared to existing models, the proposed framework significantly reduces prediction time and improves prediction accuracy.
For example, TNO reduced the average relative error by 85.5\% compared to FNO networks.
The panoramic framework performs well in various downstream tasks and can achieve results that traditional methods cannot in local sensitivity analysis.
This method improves the efficiency of both single-objective and multi-objective task optimization by 4 orders of magnitude.
\par
This study only considered changes in blade geometry. Future research could include adjusting operating boundary conditions to more comprehensively cover turbomachinery design scenarios.
Additionally, future research will explore combining active learning to focus data distribution on the most critical areas, aiming to further reduce the number of training samples.

\vspace{-0.3cm}
\section*{Conflict of interest}
The authors declare that they have no conflict of interest.
\section*{Data availability statement}
The data that support the findings of this study are available from the corresponding author upon reasonable request.
\clearpage
\section{Appendix}
\begin{figure}[htbp]
\centering
\includegraphics[width=0.75\linewidth]{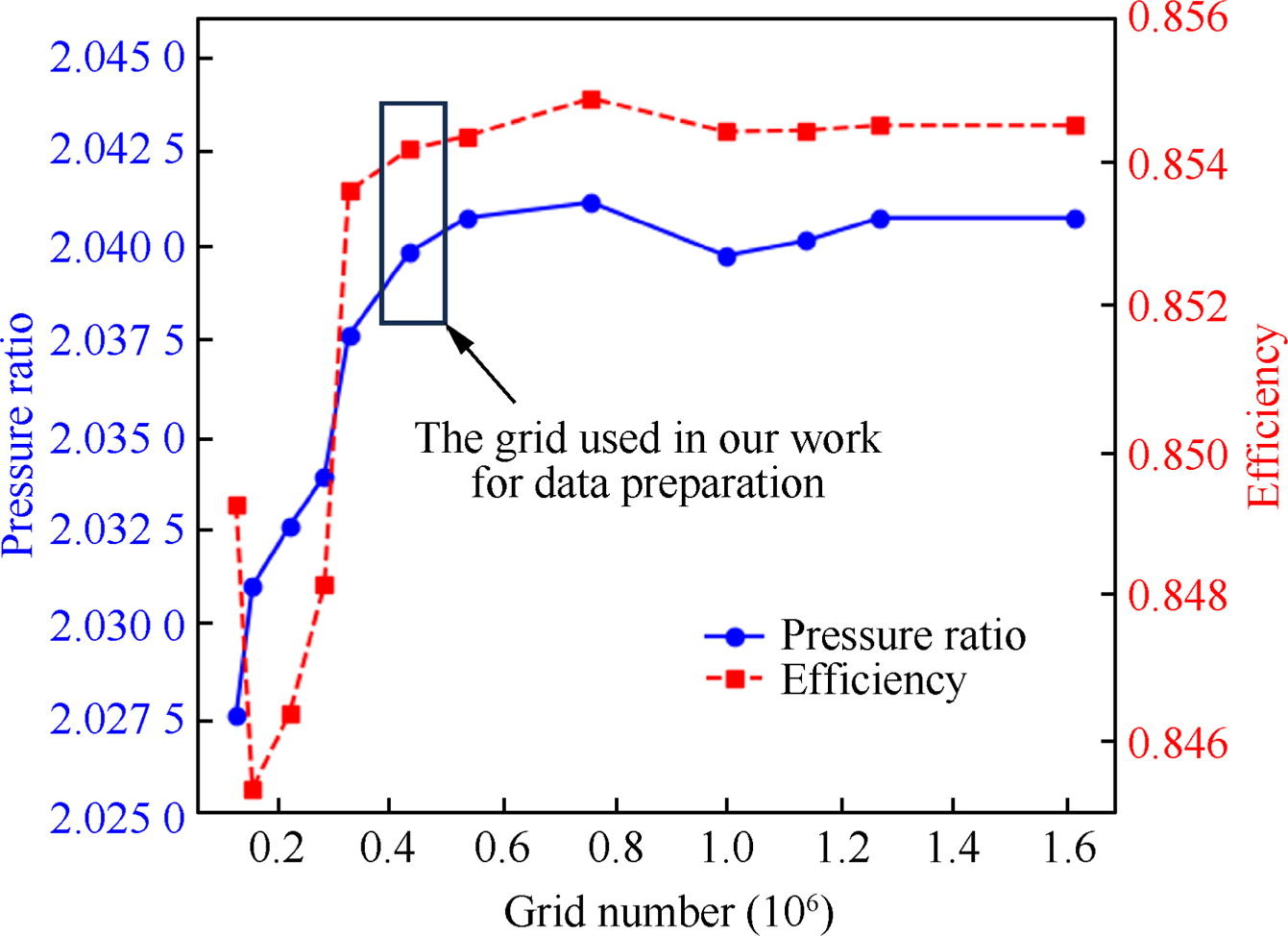}
\caption{Grid-independence verification; the black box denotes the selected grid. Reproduced from Fig. A1 of the published article under CC BY 4.0 (see title note).}
\label{grid_independence}
\end{figure}
The detailed structures of the other four deep neural networks for comparison are shown in the appendix.
  \begin{table}[htbp]
    \centering
  \caption{The detailed network structure of MLP}
  \begin{tabular}{cc}
  \toprule
  \tabincell{c}{Network type} & \tabincell{c}{Output size}  \\ \midrule
      Input & 28  \\
      (Linear+ BN1d+ GELU)$\times$5 & 256  \\
      Linear-7 & 4096$\times$5  \\ \midrule
      \multicolumn{2}{c}{Forward pass size: 768.00MB
      Params size: 21.86MB}
      \\\bottomrule
  \end{tabular}
\end{table}
\begin{table}[htbp]
  \centering
  \caption{The detailed network structure of UNet}
  \begin{tabular}{cc}
  \toprule
      Network type & \tabincell{c}{Output size} \\ \midrule
      Input & 4096$\times$(28+2)  \\
      Conv2d+ ResBlock & 4096$\times$30  \\
      Conv2d$\times$2 & 4096$\times$64  \\
      ResBlock + MaxPool2d + Conv2d$\times$2 & 64$\times$32$\times$32  \\
      ResBlock + MaxPool2d + Conv2d$\times$2& 128$\times$16$\times$16  \\
      ResBlock+ MaxPool2d + Conv2d$\times$2& 256$\times$8$\times$8  \\
      ResBlock+ MaxPool2d + Conv2d$\times$2& 512$\times$4$\times$4  \\
      \tabincell{c}{ConvTranspose2d + (Conv2d$\times$3) + ResBlock} & 512$\times$8$\times$8  \\
      \tabincell{c}{ConvTranspose2d + (Conv2d$\times$3) + ResBlock} & 256$\times$16$\times$16  \\
      \tabincell{c}{ConvTranspose2d + (Conv2d$\times$3) + ResBlock} & 128$\times$32$\times$32  \\
      \tabincell{c}{ConvTranspose2d + (Conv2d$\times$3) + ResBlock} & 4096$\times$64  \\
      Conv2d + ResBlock & 4096$\times$5  \\
      Interp2dUpsample + Conv2d & 4096$\times$5  \\ \midrule
      \multicolumn{2}{c}
      {Forward pass size: 624.56MB
      Params size: 130.41MB}
      \\\bottomrule
  \end{tabular}
\end{table}
\begin{table}[htbp]
  \centering
  \caption{The detailed network structure of deepONet}
  \begin{tabular}{cc}
  \toprule
      Branch Network type & \tabincell{c}{Output size}  \\ \midrule
      Input & 28  \\
      (Linear+ FcnSingle)$\times$4 & 128  \\
      Linear & 5 \\ \midrule
      Trunk Network type & Output size  \\ \midrule
      Input & 4096$\times$2  \\
      (Linear + GELU)$\times$10 & 4096$\times$256  \\
      Linear+ FcnSingle & 4096$\times$128  \\
      Linear & 4096$\times$5 \\ \midrule
      \multicolumn{2}{c}
      {Forward pass size: 822.56MB
      Params size: 1.79MB}
      \\\bottomrule
  \end{tabular}
\end{table}
\begin{table}[htbp]
  \centering
  \caption{The detailed network structure of FNO}
  \begin{tabular}{cc}
  \toprule
  \tabincell{c}{Network  type} & \tabincell{c}{Output size}  \\ \midrule
      Input & 4096$\times$(28+2)  \\
      Linear-1 & 4096$\times$128  \\
      (Conv2d + Dropout + GELU + SpectralConv2d)$\times$4 & 128$\times$72$\times$72  \\
      Linear-2 & 4096$\times$128  \\
      Linear-3 & 4096$\times$5  \\ \midrule
      \multicolumn{2}{c}
      {Forward pass size: 558.91MB
      Params size: 1.66MB}
      \\\bottomrule
  \end{tabular}
\end{table}
\begin{figure}[htbp]
\centering
\includegraphics[width=0.98\linewidth,height=0.83\textheight,keepaspectratio]{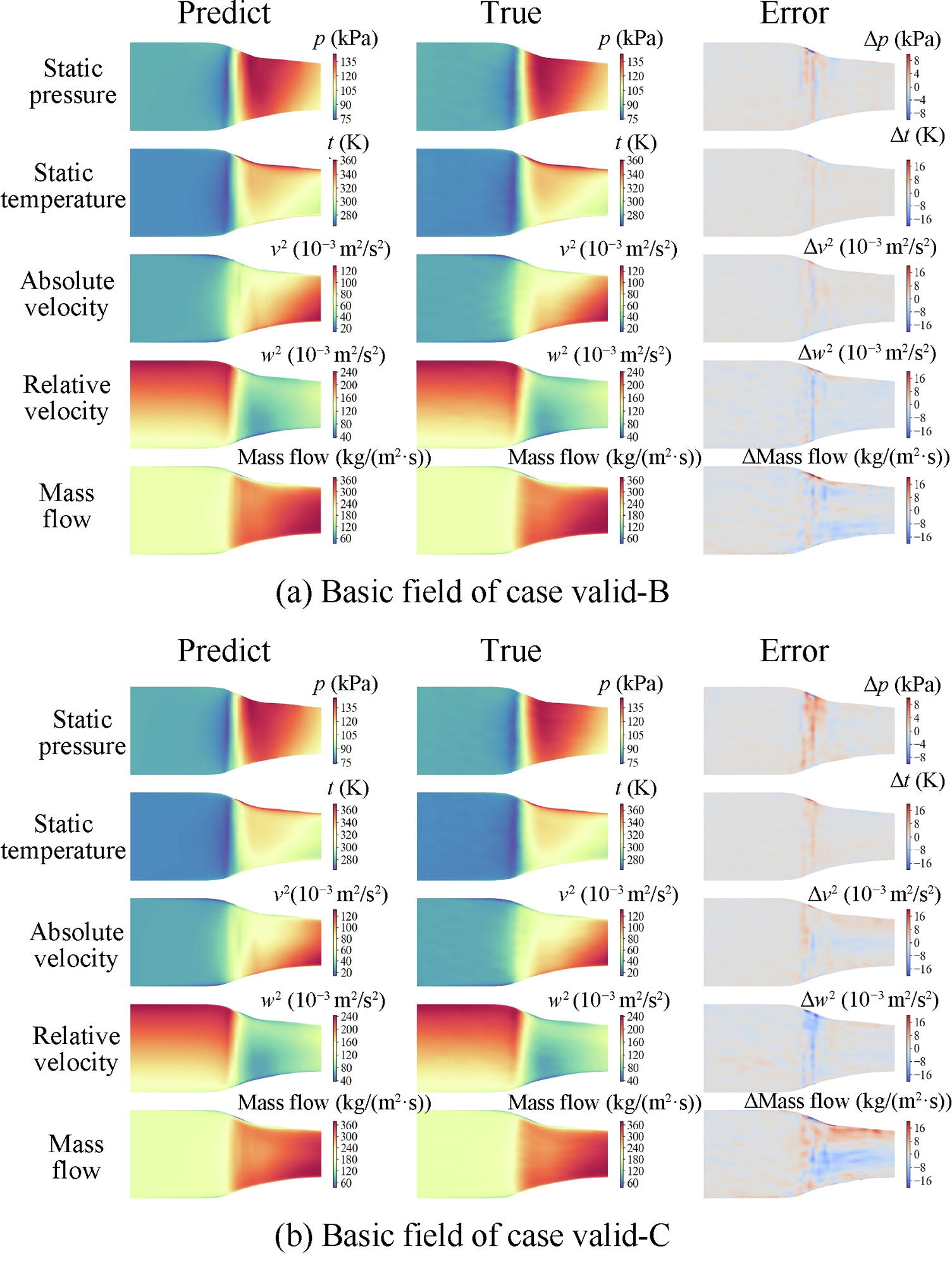}
\caption{Predicted, CFD, and error maps of basic fields for cases valid-B and valid-C. Reproduced from the upper panels of Fig. A2 of the published article under CC BY 4.0 (see title note).}
\label{fundamental_field_B}
\end{figure}
\begin{figure}[htbp]
\centering
\includegraphics[width=0.98\linewidth,height=0.83\textheight,keepaspectratio]{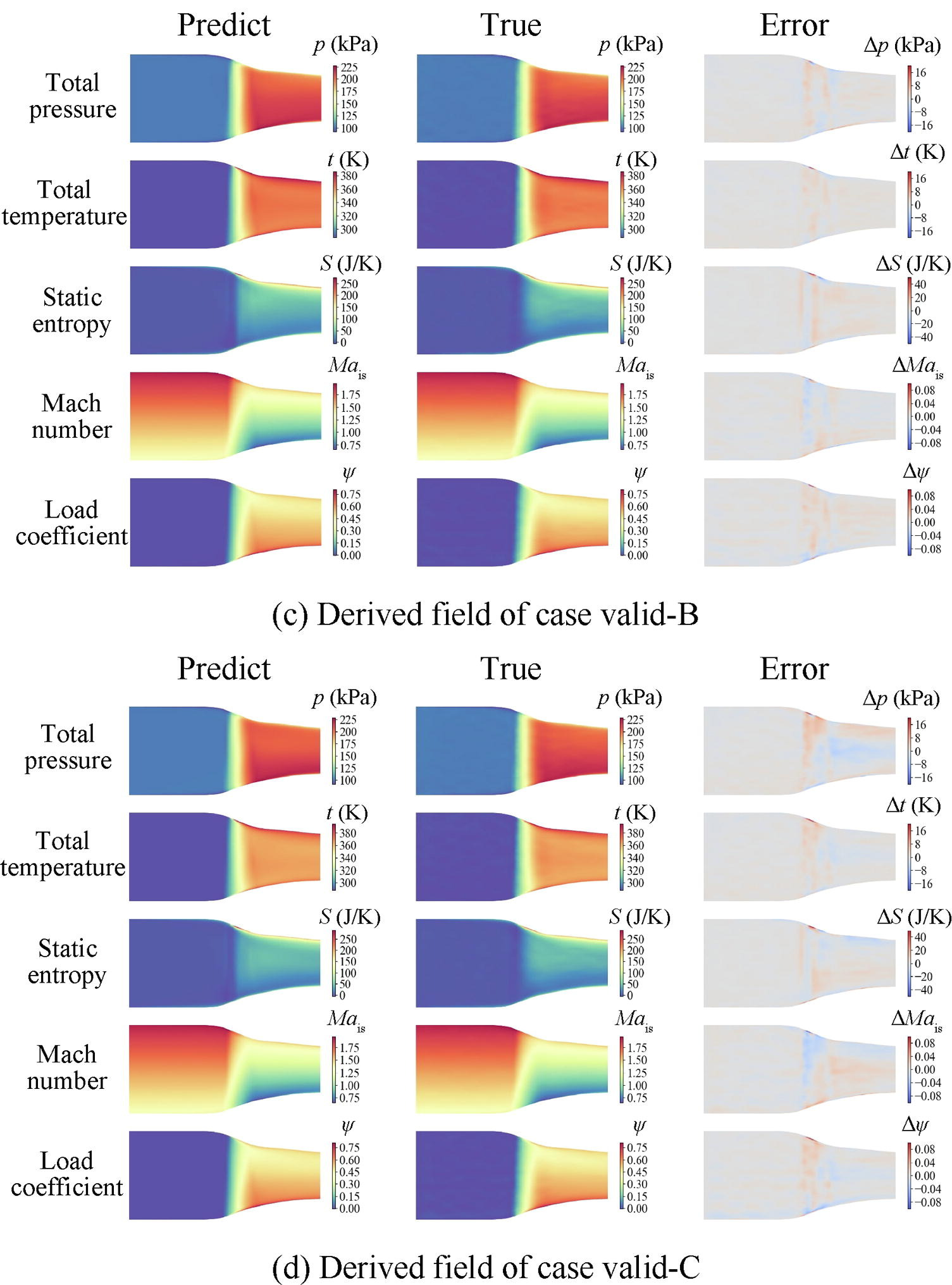}
\caption{Predicted, CFD, and error maps of derived fields for cases valid-B and valid-C. Reproduced from the lower panels of Fig. A2 of the published article under CC BY 4.0 (see title note).}
\label{derived_field_BC}
\end{figure}
\clearpage
\bibliographystyle{elsarticle-num-names}
\bibliography{bibfile}%

\end{document}